\documentclass[letterpaper]{article} % DO NOT CHANGE THIS
\usepackage{aaai25}  % DO NOT CHANGE THIS
\usepackage{times}  % DO NOT CHANGE THIS
\usepackage{helvet}  % DO NOT CHANGE THIS
\usepackage{courier}  % DO NOT CHANGE THIS
\usepackage[hyphens]{url}  % DO NOT CHANGE THIS
\usepackage{graphicx} % DO NOT CHANGE THIS
\usepackage{natbib}  % DO NOT CHANGE THIS AND DO NOT ADD ANY OPTIONS TO IT
\usepackage{caption} % DO NOT CHANGE THIS AND DO NOT ADD ANY OPTIONS TO IT
\usepackage{amsfonts}
\usepackage{amsmath}

\usepackage{algorithm}
\usepackage{algorithmicx}
\usepackage{algpseudocode} 

\usepackage{booktabs}       % professional-quality tables
\usepackage{xcolor}         % colors
\usepackage{multirow}
\usepackage{threeparttable}
\newcommand{\update}[1]{{\textcolor{black}{#1}}}
\newcommand{\boldres}[1]{{\textbf{\textcolor{red}{#1}}}}
\newcommand{\secondres}[1]{{\underline{\textcolor{blue}{#1}}}}

\usepackage{newfloat}
\usepackage{listings}
\DeclareCaptionStyle{ruled}{labelfont=normalfont,labelsep=colon,strut=off} % DO NOT CHANGE THIS
\floatstyle{ruled}
\newfloat{listing}{tb}{lst}{}
\floatname{listing}{Listing}
\nocopyright %%%%%

\title{FMamba: Mamba based on Fast-attention for Multivariate Time-series Forecasting}
\author{
    Shusen Ma\textsuperscript{\rm 1}, 
    Yu Kang\textsuperscript{\rm 1, \rm 2, \rm 3}, 
    Peng Bai\textsuperscript{\rm 2},
    Yun-Bo Zhao\textsuperscript{\rm 1, \rm 2, \rm 3}\thanks{Corresponding author.}
}
\affiliations{
    \textsuperscript{\rm 1}Institute of Advanced Technology, USTC\\
    \textsuperscript{\rm 2}Department of Automation, USTC\\
    \textsuperscript{\rm 3}Institute of Artificial Intelligence, Hefei Comprehensive National Science Center\\
    ybzhao@ustc.edu.cn\\
}

\usepackage{bibentry}
\begin{document}

\maketitle

\begin{abstract}
In multivariate time-series forecasting (MTSF), extracting the temporal correlations of the input sequences is crucial. While popular Transformer-based predictive models can perform well, their quadratic computational complexity results in inefficiency and high overhead. The recently emerged Mamba, a selective state space model, has shown promising results in many fields due to its strong temporal feature extraction capabilities and linear computational complexity. However, due to the unilateral nature of Mamba, channel-independent predictive models based on Mamba cannot attend to the relationships among all variables in the manner of Transformer-based models. To address this issue, we combine fast-attention with Mamba to introduce a novel framework named FMamba for MTSF. Technically, we first extract the temporal features of the input variables through an embedding layer, then compute the dependencies among input variables via the fast-attention module. Subsequently, we use Mamba to selectively deal with the input features and further extract the temporal dependencies of the variables through the multi-layer perceptron block (MLP-block). Finally, FMamba obtains the predictive results through the projector, a linear layer. Experimental results on eight public datasets demonstrate that FMamba can achieve state-of-the-art performance while maintaining low computational overhead.
\end{abstract}

\section{Introduction}

\begin{figure*}[ht]
	\centering
	\includegraphics[width=0.9\textwidth]{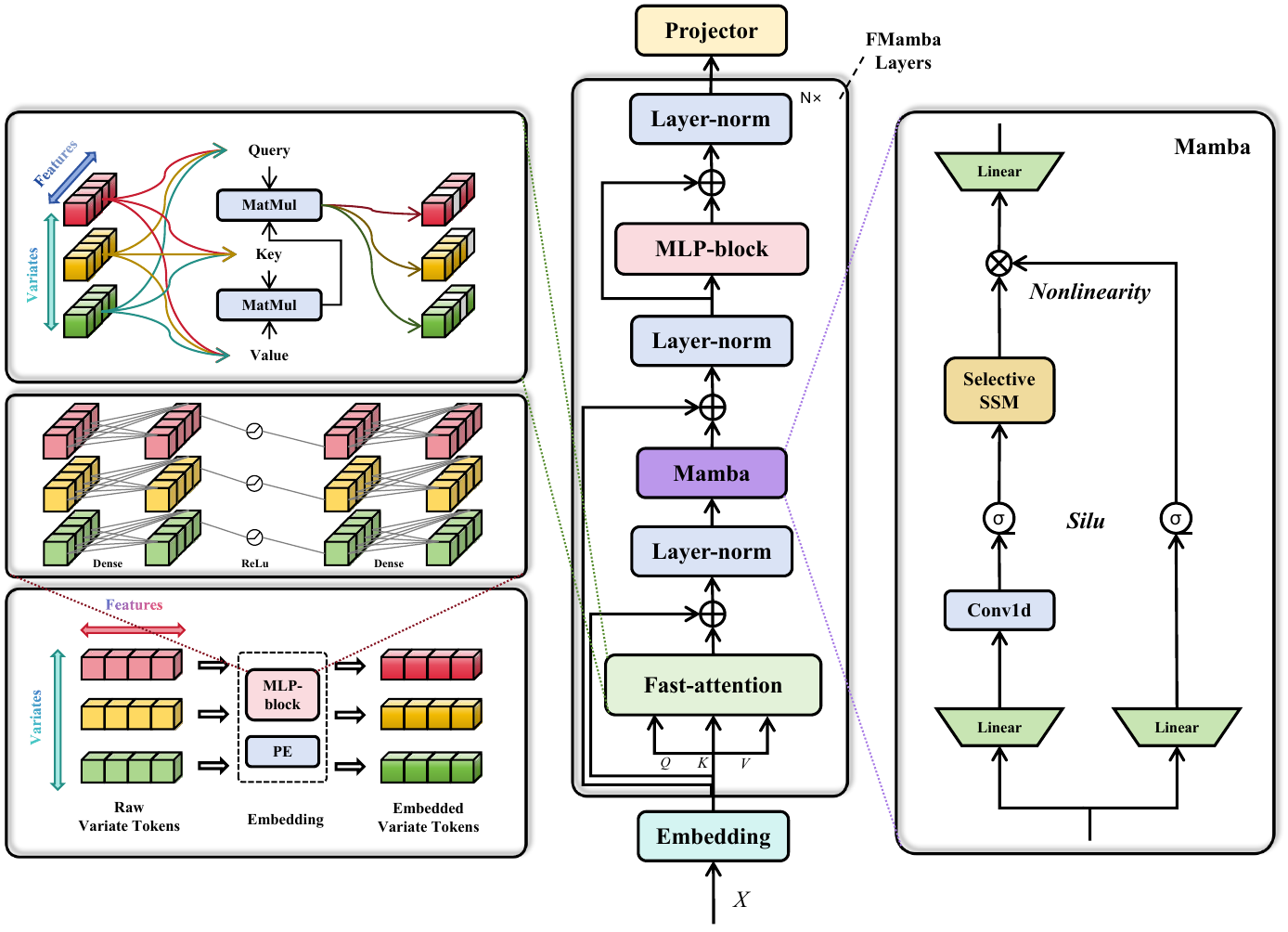}
	\caption{The structure of FMamba.}
	\label{model}
\end{figure*}

Multivariate time-series forecasting (MTSF) tasks involve discerning inter-variable correlations and intra-series temporal dependencies by learning the historical observations of diverse variables. The learned prior knowledge is then used to predict the future states of these variables over a certain period. Inter-variable correlations refer to the correlations between different variables, such as how traffic flow might decrease on rainy days because people perceive it as more dangerous to travel. Intra-series temporal dependencies refer to the dependencies between different time points within a sequence. For example, if the temperature is high on a particular day, it might tend to fluctuate around that value in the following days. Learning the inter-variable correlations and intra-series dependencies can help the model make better predictions about the future states of the variables \cite{liu2024itransformer, 10552140}.

Currently, the most popular models for MTSF are those based on Transformers, with most adopting a channel-mixing approach \cite{zhou2022fedformer, zhang2022crossformer}. However, due to the quadratic computational complexity of self-attention mechanisms, these models' computational cost and efficiency increase and decrease dramatically with the increasing input length. Although DLinear \cite{zeng2023transformers} demonstrates that simple models based on linear networks could outperform Transformer-based models in prediction performance, subsequent studies, such as PatchTST \cite{Yuqietal-2023-PatchTST} and iTransformer \cite{liu2024itransformer}, reaffirm the effectiveness of Transformer-based models by adopting channel-independent data processing methods, despite not fundamentally solve the issue of quadratic computational complexity. The model proposed in this paper is also built on the foundation of input channel independence.

To find a model structure that can replace the Transformer, Gu et al. have been promoting the development of State Space Models (SSM) \cite{gu2021efficiently}. SSM can be regarded as a combination of convolutional neural network (CNN) and recurrent neural network (RNN), achieving linear computational efficiency. However, SSM cannot dynamically adjust internal parameters based on different inputs, making it inefficient at extracting temporal features from input sequences. To address this issue, Mamba \cite{gu2023mamba} parameterizes the input of SSM to enable selective processing of input sequences. Nevertheless, due to the unilateral nature of Mamba, it cannot attend to the global variable correlations in the manner of the self-attention mechanism. 

To solve the above problem, we apply a fast-attention mechanism to the Mamba, introducing a new prediction model with linear computational complexity called FMamba shown in Figure~\ref{model}. In FMamba, the input sequence first passes through an Embedding layer to preliminarily extract temporal information. Then, the fast-attention mechanism \cite{ChoromanskiLDSG21} efficiently captures the correlations between diverse variables. After that, Mamba parameterizes the input features, allowing FMamba to focus on or ignore information between variables selectively. Subsequently, a multi-layer perceptron block (MLP-block) module extracts deeper temporal information. Finally, a linear layer, called the projector, generates the outcome. Experimentally, the proposed FMamba has demonstrated its effectiveness across various real-world forecasting benchmarks compared with the current state-of-the-art (SOTA) model. The main contributions of this paper are summarized as follows:
\begin{itemize}
    \item We propose a novel and simple prediction framework called FMamba, which achieves SOTA-level performance and obtains linear computational complexity, significantly reducing the computational overhead and enhancing the model's computational efficiency.
    \item We innovatively combine the fast-attention mechanism with the Mamba. The fast-attention mechanism can prevent the model from failing to attend to the correlations of global variables due to the unilateral nature of Mamba, while the introduction of Mamba helps the model better focus on valuable information between variables.
    \item Experimental results demonstrate that the proposed FMamba achieves SOTA performance on eight public datasets.
\end{itemize}

\section{Related Work}

The traditional MTSF models are based on statistical methods, such as AR (autoregression), MA (moving average), ARMA (autoregression moving average), and ARIMA (autoregression integrated moving average). Although highly interpretable, these methods cannot effectively handle complex nonlinear relationships between various variables. On the other hand, due to the robust nonlinear fitting capability of deep neural networks (DNN), prediction models based on DNN have excelled in MTSF tasks, with classic models being based on RNN. However, the cyclic structure of RNN makes it unsuitable for processing long sequences as it may encounter gradient explosion or gradient vanishing issues. While variants like long short-term memory (LSTM) and gated recurrent unit (GRU) mitigate these problems to some extent through gating mechanisms, they still retain the cyclic structure. Another class of predictive models is based on CNN, which utilizes local perception capabilities to capture the temporal characteristics of variables. Additionally, hybrid models \cite{wang2023hybrid, ma2023tcln} are also introduced widely.

With the booming development of Transformer \cite{NIPS2017_3f5ee243} in natural language processing and computer vision, many Transformer-based models have been proposed, such as FEDformer \cite{zhou2022fedformer} and Crossformer \cite{zhang2022crossformer}. However, recent studies \cite{zeng2023transformers, das2023long} have questioned the necessity of Transformer-based models for MTSF, as simple linear layers can achieve or even surpass the performance of Transformer-based models. To address these concerns, models like iTransformer \cite{liu2024itransformer} and PatchTST \cite{Yuqietal-2023-PatchTST} have been proposed. iTransformer treats each input variable as a token, a method also adopted by PCDformer \cite{10552140}, and then uses the self-attention mechanism to calculate the correlations between variables. PatchTST divides the input data into diverse channels, allowing different variables to share the same Transformer backbone. Experiments have shown that this new data processing method enables Transformer-based models to once again achieve SOTA-level performance.

However, the self-attention mechanism of Transformer models results in quadratic computational complexity. To find a better framework, SSM has been proposed, reducing computational costs while maintaining strong long-term temporal dependency extraction capabilities. Mamba introduces a selective mechanism based on SSM, enabling the model to selectively focus on or ignore certain inputs during inference through parameterizing the input. TimeMachine \cite{ahamed2024timemachine} is the first to leverage purely SSM modules to capture long-term dependencies in MTSF tasks, employing channel-mixing and channel-independent strategies. To address the limitation of Mamba in MTSF tasks, where it cannot attend to global variables due to its unilateral nature, S-Mamba \cite{wang2024mamba} reconfigures Mamba blocks for bidirectional scanning. However, this approach cannot directly calculate the correlation between different variables, as the self-attention mechanism does. To better capture the correlations between variables and the temporal features within sequences, this paper attempts to combine fast-attention and Mamba. By leveraging the advantages of both fast-attention and Mamba while maintaining linear computational complexity, we aim to create a novel, simple, and effective prediction framework for MTSF tasks.

\section{Methodology}

In this section, we first briefly explain the problem statement and then provide a detailed introduction to the structure of FMamba.

\subsection{Problem Statement}
Consider a dataset $\mathbf{X} \in \mathbb{R}^{l \times n}$, where $l$ denotes the total length of the dataset and $n$ denotes the number of variables. Given a fixed-length look-back window $L$, the objective of the MTSF task is to predict the future sequence $\widetilde{\mathbf{X}}_{t+1:t+\tau} = \{ \mathbf{x}_{t+1}, \ldots, \mathbf{x}_{t+\tau} \}$, leveraging historical observations $\mathbf{X}_{t-L+1:t} = \{ \mathbf{x}_{t-L+1}, \ldots, \mathbf{x}_{t} \}$. Here, $\tau$ denotes the prediction horizon, and $\mathbf{x_t} \in \mathbb{R}^{n}$ signifies the system's state at time $t$. 
In this work, we adopt a channel-independent approach \cite{10552140, liu2024itransformer} to process the input sequence. Therefore, the input for FMamba can be represented as $ \mathbf{X} = \{ \mathbf{X}_{\rm token} , \mathbf{X}_0  \} \in \mathbb{R}^{ n \times ( L+\tau ) } $, where $\mathbf{X}_{\rm token} \in \mathbb{R}^{ n \times L } $ and $\mathbf{X}_0 \in \mathbb{R}^{ n \times \tau }$ that is initialized by zero \cite{zhou2021informer}. Channel independence means that different input variables correspond to various channels.

\subsection{Embedding Layer}
The embedding layer comprises positional embedding (PE) \cite{NIPS2017_3f5ee243} and an MLP-block. PE aims to add positional information for each time step, aiding the model in understanding the temporal relationships within sequential data. The purpose of the MLP-block is to transform the input of each variable into a high-dimensional representation to capture temporal features. The specific computation formula is as follows:
\begin{equation}
    \text{MLP-block}(x) = \text{ReLU}(\mathbf{W}_1 x + \mathbf{b}_1) \mathbf{W}_2 + \mathbf{b}_2,
\end{equation}
where $\mathbf{W}_1, \mathbf{W}_2$ are weight matrices, and $\mathbf{b}_1, \mathbf{b}_2$ are bias vectors. The final output of the embedding layer is the sum of the outputs of PE and MLP-block.

\subsection{Fast-attention Mechanism}

\begin{figure}[ht]
	\centering
	\includegraphics[width=0.48\textwidth]{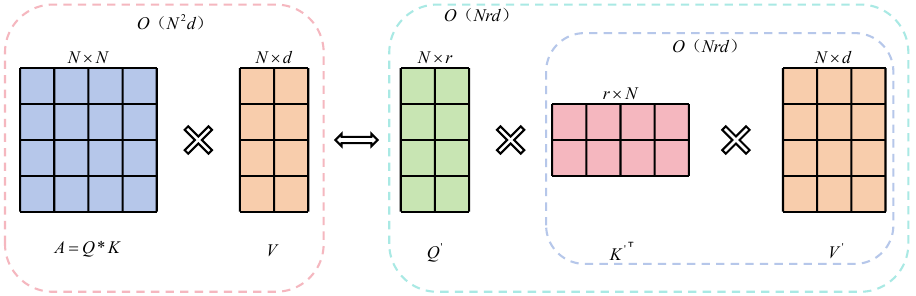}
	\caption{The illustration of canonical self-attention and fast-attention.}
	\label{fast-attention}
\end{figure}

Canonical self-attention shown on the left side of Figure \ref{fast-attention}, also known as scaled dot-product attention, is a mechanism that allows a model to focus on different parts of an input sequence when computing a representation for each element in the sequence. The specific computation formula is as follows:
\begin{equation}
\begin{aligned}
&\mathbf{Q} = \mathbf{X} \mathbf{W}_Q , \quad \mathbf{K} = \mathbf{X} \mathbf{W}_K , \quad \mathbf{V} = \mathbf{X} \mathbf{W}_V, \\
&\text{Attention}(\mathbf{Q}, \mathbf{K}, \mathbf{V}) = \text{softmax}\left(\frac{\mathbf{Q} \mathbf{K}^\top}{\sqrt{d_k}}\right) \mathbf{V}.
\end{aligned}
\end{equation}
The $\mathbf{Q}$, $\mathbf{K}$, and $\mathbf{V}$ correspond to the query, key, and value matrices, respectively. The $\mathbf{W}_Q$, $\mathbf{W}_K$, and $\mathbf{W}_V$ are all learnable parameters. The $d_k$ is the dimension of the key.

However, due to its quadratic computational complexity, its computational overhead increases rapidly with the increase of the input sequence's length. Inspired by \cite{ChoromanskiLDSG21}, we attempt to use the fast-attention mechanism, shown on the right side of Figure \ref{fast-attention}, to reduce computational overhead while enabling the model to discern the correlation between input variables. In this paper, we use the Gaussian kernel function $\mathcal{F}$ to transform the input nonlinearly, obtaining the query and key. It can effectively smooth data, and capture local patterns and nonlinear relationships. Unlike self-attention, fast-attention first multiplies $\mathbf{K'}^\top$ and $\mathbf{V'}$ and then multiplies the product with $\mathbf{Q'}$, achieving linear computational complexity. The specific calculation formula is as follows:
\begin{equation}
\begin{aligned}
& \mathcal{F}(x) = \textbf{exp} (\frac{-x^2}{2}) , \\
&\mathbf{Q'} = \mathcal{F}_{\mathbf{Q'}}(\mathbf{X}), \quad \mathbf{K'} =  \mathcal{F}_{\mathbf{K'}}(\mathbf{X}), \quad \mathbf{V'} =  \mathbf{X} \mathbf{W}_{V'} , \\
&\text{Attention}(\mathbf{Q'}, \mathbf{K'}, \mathbf{V'}) = \frac{\mathbf{Q'}}{k\_{dim}}  \left(\mathbf{K'}^\top \mathbf{V'}\right),
\end{aligned}
\end{equation}
where $k\_{dim}$ denotes the kernel dimension of the fast-attention.

\subsection{Mamba}

The SSM represents the internal state evolution of the system through first-order differential equations and controls the system's output through latent state representations, defined as follows:
\begin{equation}
\begin{aligned}
h(t)^{\prime} & =\boldsymbol{A} h(t)+\boldsymbol{B} x(t)  ,\\
y(t) & =\boldsymbol{C} h(t),
\end{aligned}
\end{equation}
where $h(t)$ represents the latent state representation at any given time $t$, and $x(t)$ represents the input at time $t$. $\boldsymbol{A} \in \mathbb{R}^{N \times N}$, $\boldsymbol{B} \in \mathbb{R}^{N \times D}$, and $\boldsymbol{C} \in \mathbb{R}^{N \times D}$ are all learnable parameter matrices. In real-world applications, especially in MTSF tasks, the data are usually discrete time series. Therefore, the model needs to be discretized to accommodate these discrete-time data. The following formula illustrates how to transform a continuous SSM into a discrete SSM by using zero-order holding and time sampling at intervals of $\Delta$:
\begin{equation}
\begin{aligned}
& \overline{\boldsymbol{A}}=\exp (\Delta \boldsymbol{A}), \\
& \overline{\boldsymbol{B}}=(\Delta \boldsymbol{A})^{-1}(\exp (\Delta \boldsymbol{A})-\boldsymbol{I}) \cdot \Delta \boldsymbol{B}  ,\\
& h_k=\overline{\boldsymbol{A}} h_{k-1}+\overline{\boldsymbol{B}} x_k  ,\\
& y_k=\boldsymbol{C} h_k.
\end{aligned}
\end{equation}

As a parameterized mapping from input data to output data, SSM can be seen as a combination of CNN and RNN. Specifically, a CNN structure is used during training, and an RNN structure is used during inference. Since $\overline{\boldsymbol{A}}$ only remembers the previously captured state information, Gu et al. introduced HiPPO \cite{NEURIPS2020_102f0bb6} to efficiently solve long-range dependency problems in sequence modeling with limited storage space. This constitutes the core of the structured state space model (S4) \cite{gu2021efficiently}.

However, there is a problem with SSM: the matrices $\boldsymbol{A}$, $\boldsymbol{B}$, and $\boldsymbol{C}$ do not vary with diverse inputs, making it impossible to perform targeted inference for various inputs. To address this issue, Mamba \cite{gu2023mamba} designed a simple selection mechanism that parameterizes the input of SSM, allowing the model to selectively focus on or ignore input information during inference. Moreover, Mamba employs a hardware-aware parallel algorithm to ensure computational efficiency. The structure of Mamba is shown on the right side of Figure \ref{model}, and its specific process is illustrated in Algorithm \ref{process_of_mamba_block}. The input sequence received by Mamba is $X \in \mathbb{R}^{B \times V \times D}$, where $B$ represents the batch size, $V$ represents the number of input variables, and $D$ represents the feature representation dimension. Similar to the Transformer, the output dimension of the Mamba remains consistent with the input dimension.

\begin{algorithm}
\caption{The process of Mamba Block}
\label{process_of_mamba_block}
{\bf Input:} $X: (B, V, D)$

{\bf Output:} $Y: (B, V, D)$
\begin{algorithmic}[1]
\State $x: (B, V, E) \leftarrow \texttt{Linear}_x(X) $ \Comment{Linear projection}
\State $z : (B, V, E) \leftarrow \texttt{Linear}_z(X) $
\State $x' : (B, V, E) \leftarrow \texttt{SiLU}(\texttt{Conv1D}(x))$
\State $\boldsymbol{A} : (D, N) \leftarrow \texttt{Parameter} $ \Comment{Structured state matrix}
\State $\boldsymbol{B}: (B, V, N) \leftarrow \texttt{Linear}_B(x')$
\State $\boldsymbol{C} : (B, V, N) \leftarrow \texttt{Linear}_C(x')$
\State $\Delta : (B, V, D) \leftarrow \texttt{Softplus}(\texttt{Parameter} + \texttt{Broadcast}(\texttt{Linear}(x')))$
\State $\overline{\boldsymbol{A}}, \overline{\boldsymbol{B}} : (B, V, D, N) \leftarrow \texttt{discretize}(\Delta, \boldsymbol{A}, \boldsymbol{B}) $ \Comment{Input-dependent parameters and discretization}
\State $y : (B, V, E) \leftarrow \texttt{SelectiveSSM}(\overline{\boldsymbol{A}}, \overline{\boldsymbol{B}}, \boldsymbol{C})(x')$
\State $y' : (B, V, E) \leftarrow y \otimes \texttt{SiLU}(z)$
\State $Y : (B, V, D) \leftarrow \texttt{Linear}(y') $
\end{algorithmic}
\end{algorithm}

Specifically, the input of Mamba undergoes linear projections to generate $x$ and $z$ respectively. The further one-dimensional convolution operation is applied to $x$, followed by a SiLU activation function to produce $x'$. $\boldsymbol{A}$ is initialized and represents the structured state matrix. $\boldsymbol{B}$ and $\boldsymbol{C}$ are generated from $x'$ through linear transformation. $\Delta$ is computed by applying a Softplus function to a combination of the parameter and a broadcasted linear transformation of $x'$. $\overline{\boldsymbol{A}}$ and $\overline{\boldsymbol{B}}$ are produced by discretization operation.
Through the above operations, Mamba makes $\boldsymbol{B}$, $\boldsymbol{C}$, and $\Delta$ functions of the input sequence. Although $\boldsymbol{A}$ is not data-dependent, $\overline{\boldsymbol{A}}=\exp (\Delta \boldsymbol{A})$ makes it true. This transformation allows Mamba to adapt its behavior based on the input content, ultimately making Mamba a data-dependent selective model.

\section{Experimental Results}

\subsection{Datasets and Baselines}

\begin{table}[ht]
	\caption{The overall information of the eight datasets.}
	\begin{center}
		\scalebox{0.8}{
			\begin{tabular}{c|ccc}
				\hline
				Datasets       & Variants     & Dataset Size      & Granularity      \\ \hline
				PEMS03       & 358             & (15617, 5135, 5135)      & 5min     \\
				PEMS04       & 307             & (10172, 3375, 3375)      & 5min     \\
				PEMS07       & 883             & (16911, 5622, 5622)       & 5min    \\ 
                PEMS08       & 170             & (10690, 3548, 3548)       & 5min    \\ \hline
                Electricity       & 321             & (18317, 2633, 5261)      & 1hour    \\
                SML2010       & 22             & (2752, 368, 780)       & 15min     \\
                Weather       & 21             & (36792, 5271, 10540)      & 10min    \\
                Solar-Energy       & 137             & (36601, 5161, 10417)       & 10min    \\ \hline

		\end{tabular}
  }
	\end{center}
	\label{datasets}
% \vspace{-8pt}
\end{table}

The primary experiments are carried out on the subsequent eight publicly available datasets: (1) PEMS (PEMS03, PEMS04, PEMS07, and PEMS08):  containing California's public traffic network data, collected in 5-minute intervals; (2) Electricity: the hourly data on electricity consumption from 321 clients; (3) SML2010: collecting from 22 sensors in a room for about 40 days with an average sampling time of 15 minutes; (4) Weather: including 21 meteorological variables recorded every 10 minutes at the Weather Station of the Max Planck Biogeochemistry Institute in 2020; (5) Solar-Energy: recording the solar power production of 137 PV plants in 2006, which are sampled every 10 minutes. More details about the datasets are shown in Table~\ref{datasets}.

We primarily compare FMamba with 11 current SOTA models, including S-Mamba \cite{wang2024mamba}, iTransformer \cite{liu2024itransformer}, RLinear \cite{li2023revisiting}, PatchTST \cite{Yuqietal-2023-PatchTST}, Crossformer \cite{zhang2022crossformer}, TiDE \cite{das2023long}, TimesNet \cite{wu2022timesnet}, DLinear \cite{zeng2023transformers}, SCINet \cite{liu2022scinet}, FEDformer \cite{zhou2022fedformer}, and Stationary \cite{NEURIPS2022_4054556f}.

\begin{table*}[!ht]
	\caption{Full results of diverse MTSF tasks. Extensive baselines are compared under different forecasting lengths following the setting of iTransformer. The input length is set to 96 for all baselines. Avg is the average of all four prediction lengths. * denotes reimplementation and the other baselines' results refer to iTransformer.
	}\label{full_baseline_results}

	\renewcommand{\arraystretch}{0.85} 
	\centering
	\resizebox{\textwidth}{!}
	{
		\begin{threeparttable}
				\renewcommand{\multirowsetup}{\centering}
				\setlength{\tabcolsep}{1pt}
				\begin{tabular}{c|c|cc|cc|cc|cc|cc|cc|cc|cc|cc|cc|cc|cc}
					\toprule
					\multicolumn{2}{c|}{\multirow{1}{*}{Models}} &
					\multicolumn{2}{c}{\rotatebox{0}{\scalebox{0.8}{\textbf{FMamba}}}} & 
					\multicolumn{2}{c}{\rotatebox{0}{\scalebox{0.8}{S-Mamba* }}} & 
					\multicolumn{2}{c}{\rotatebox{0}{\scalebox{0.8}{iTransformer }}} &
					\multicolumn{2}{c}{\rotatebox{0}{\scalebox{0.8}{\update{RLinear}}}} &
					\multicolumn{2}{c}{\rotatebox{0}{\scalebox{0.8}{PatchTST}}} &
					\multicolumn{2}{c}{\rotatebox{0}{\scalebox{0.8}{Crossformer}}} &
					\multicolumn{2}{c}{\rotatebox{0}{\scalebox{0.8}{TiDE}}} &
					\multicolumn{2}{c}{\rotatebox{0}{\scalebox{0.8}{{TimesNet}}}} &
					\multicolumn{2}{c}{\rotatebox{0}{\scalebox{0.8}{DLinear}}} &
					\multicolumn{2}{c}{\rotatebox{0}{\scalebox{0.8}{SCINet}}} &
					\multicolumn{2}{c}{\rotatebox{0}{\scalebox{0.8}{FEDformer}}} &
					\multicolumn{2}{c}{\rotatebox{0}{\scalebox{0.8}{Stationary}}}  \\

					\cmidrule(lr){3-4} \cmidrule(lr){5-6}\cmidrule(lr){7-8} \cmidrule(lr){9-10}\cmidrule(lr){11-12}\cmidrule(lr){13-14} \cmidrule(lr){15-16} \cmidrule(lr){17-18} \cmidrule(lr){19-20} \cmidrule(lr){21-22} \cmidrule(lr){23-24} \cmidrule(lr){25-26}
					\multicolumn{2}{c|}{Metric}  & \scalebox{0.78}{MSE} & \scalebox{0.78}{MAE}  & \scalebox{0.78}{MSE} & \scalebox{0.78}{MAE}  & \scalebox{0.78}{MSE} & \scalebox{0.78}{MAE}  & \scalebox{0.78}{MSE} & \scalebox{0.78}{MAE}  & \scalebox{0.78}{MSE} & \scalebox{0.78}{MAE}  & \scalebox{0.78}{MSE} & \scalebox{0.78}{MAE} & \scalebox{0.78}{MSE} & \scalebox{0.78}{MAE} & \scalebox{0.78}{MSE} & \scalebox{0.78}{MAE} & \scalebox{0.78}{MSE} & \scalebox{0.78}{MAE} & \scalebox{0.78}{MSE} & \scalebox{0.78}{MAE} & \scalebox{0.78}{MSE} & \scalebox{0.78}{MAE} & \scalebox{0.78}{MSE} & \scalebox{0.78}{MAE} \\
					\toprule
					
					\multirow{5}{*}{\rotatebox{90}{\scalebox{0.95}{PEMS03}}}
					&  \scalebox{0.78}{12} & \boldres{\scalebox{0.78}{0.063}} &\boldres{\scalebox{0.78}{0.166}} &\secondres{\scalebox{0.78}{0.066}} &\secondres{\scalebox{0.78}{0.170}} & {\scalebox{0.78}{0.071}} &{\scalebox{0.78}{0.174}} &\scalebox{0.78}{0.126} &\scalebox{0.78}{0.236} &\scalebox{0.78}{0.099} &\scalebox{0.78}{0.216} &\scalebox{0.78}{0.090} &\scalebox{0.78}{0.203} & \scalebox{0.78}{0.178} & \scalebox{0.78}{0.305} &\scalebox{0.78}{0.085} &\scalebox{0.78}{0.192} &\scalebox{0.78}{0.122} &\scalebox{0.78}{0.243} &\secondres{\scalebox{0.78}{0.066}} &{\scalebox{0.78}{0.172}} &\scalebox{0.78}{0.126} &\scalebox{0.78}{0.251} &\scalebox{0.78}{0.081} &\scalebox{0.78}{0.188}     \\% &\scalebox{0.78}{0.126} &\scalebox{0.78}{0.233}\\
     
					& \scalebox{0.78}{24} & \boldres{\scalebox{0.78}{0.080}} &\boldres{\scalebox{0.78}{0.189}} &{\scalebox{0.78}{0.088}} &\secondres{\scalebox{0.78}{0.197}} &{\scalebox{0.78}{0.093}} &{\scalebox{0.78}{0.201}} &\scalebox{0.78}{0.246} &\scalebox{0.78}{0.334} &\scalebox{0.78}{0.142} &\scalebox{0.78}{0.259} &\scalebox{0.78}{0.121} &\scalebox{0.78}{0.240} & \scalebox{0.78}{0.257} & \scalebox{0.78}{0.371} &\scalebox{0.78}{0.118} &\scalebox{0.78}{0.223} &\scalebox{0.78}{0.201} &\scalebox{0.78}{0.317} &\secondres{\scalebox{0.78}{0.085}} &{\scalebox{0.78}{0.198}} &\scalebox{0.78}{0.149} &\scalebox{0.78}{0.275} &\scalebox{0.78}{0.105} &\scalebox{0.78}{0.214}  \\%&\scalebox{0.78}{0.139} &\scalebox{0.78}{0.250}\\
     
					& \scalebox{0.78}{48} & \boldres{\scalebox{0.78}{0.111}} &\boldres{\scalebox{0.78}{0.224}} &{\scalebox{0.78}{0.165}} &{\scalebox{0.78}{0.277}} &\secondres{\scalebox{0.78}{0.125}} &\secondres{\scalebox{0.78}{0.236}} &\scalebox{0.78}{0.551} &\scalebox{0.78}{0.529} &\scalebox{0.78}{0.211} &\scalebox{0.78}{0.319}  &\scalebox{0.78}{0.202} &\scalebox{0.78}{0.317} & \scalebox{0.78}{0.379}& \scalebox{0.78}{0.463} &\scalebox{0.78}{0.155} &\scalebox{0.78}{0.260} &\scalebox{0.78}{0.333} &\scalebox{0.78}{0.425} &{\scalebox{0.78}{0.127}} &{\scalebox{0.78}{0.238}} &\scalebox{0.78}{0.227} &\scalebox{0.78}{0.348} &\scalebox{0.78}{0.154} &\scalebox{0.78}{0.257}  \\% &\scalebox{0.78}{0.186} &\scalebox{0.78}{0.289}\\
     
					& \scalebox{0.78}{96} & \secondres{\scalebox{0.78}{0.166}} &\secondres{\scalebox{0.78}{0.279}} &{\scalebox{0.78}{0.226}} &{\scalebox{0.78}{0.321}} &\boldres{\scalebox{0.78}{0.164}} &\boldres{\scalebox{0.78}{0.275}} &\scalebox{0.78}{1.057} &\scalebox{0.78}{0.787} &\scalebox{0.78}{0.269} &\scalebox{0.78}{0.370} &\scalebox{0.78}{0.262} &\scalebox{0.78}{0.367} & \scalebox{0.78}{0.490}& \scalebox{0.78}{0.539} &\scalebox{0.78}{0.228} &\scalebox{0.78}{0.317} &\scalebox{0.78}{0.457} &\scalebox{0.78}{0.515} &{\scalebox{0.78}{0.178}} &{\scalebox{0.78}{0.287}} &\scalebox{0.78}{0.348} &\scalebox{0.78}{0.434} &\scalebox{0.78}{0.247} &\scalebox{0.78}{0.336} \\% &\scalebox{0.78}{0.233} &\scalebox{0.78}{0.323}\\
     
					\cmidrule(lr){2-24}
					& \scalebox{0.78}{Avg} & \boldres{\scalebox{0.78}{0.105}} &\boldres{\scalebox{0.78}{0.215}} &{\scalebox{0.78}{0.136}} &{\scalebox{0.78}{0.241}} &\secondres{\scalebox{0.78}{0.113}} &\secondres{\scalebox{0.78}{0.221}} &\scalebox{0.78}{0.495} &\scalebox{0.78}{0.472} &\scalebox{0.78}{0.180} &\scalebox{0.78}{0.291} &\scalebox{0.78}{0.169} &\scalebox{0.78}{0.281} & \scalebox{0.78}{0.326}& \scalebox{0.78}{0.419} &\scalebox{0.78}{0.147} &\scalebox{0.78}{0.248} &\scalebox{0.78}{0.278} &\scalebox{0.78}{0.375} &{\scalebox{0.78}{0.114}} &{\scalebox{0.78}{0.224}} &\scalebox{0.78}{0.213} &\scalebox{0.78}{0.327} &\scalebox{0.78}{0.147} &\scalebox{0.78}{0.249} \\% &\scalebox{0.78}{0.171} &\scalebox{0.78}{0.274}\\
					
					\midrule
					\multirow{5}{*}{\update{\rotatebox{90}{\scalebox{0.95}{PEMS04}}}} 
					&  \scalebox{0.78}{12} & \boldres{\scalebox{0.78}{0.071}} &\boldres{\scalebox{0.78}{0.175}} &\secondres{\scalebox{0.78}{0.072}} &\secondres{\scalebox{0.78}{0.177}} &{\scalebox{0.78}{0.078}} &{\scalebox{0.78}{0.183}} &\scalebox{0.78}{0.138} &\scalebox{0.78}{0.252} &\scalebox{0.78}{0.105} &\scalebox{0.78}{0.224} &\scalebox{0.78}{0.098} &\scalebox{0.78}{0.218} & \scalebox{0.78}{0.219}& \scalebox{0.78}{0.340} &\scalebox{0.78}{0.087} &\scalebox{0.78}{0.195} &\scalebox{0.78}{0.148} &\scalebox{0.78}{0.272} &{\scalebox{0.78}{0.073}} &\secondres{\scalebox{0.78}{0.177}} &\scalebox{0.78}{0.138} &\scalebox{0.78}{0.262} &\scalebox{0.78}{0.088} &\scalebox{0.78}{0.196}  \\% &\scalebox{0.78}{0.112} &\scalebox{0.78}{0.222}\\
     
					& \scalebox{0.78}{24} & \boldres{\scalebox{0.78}{0.082}} &\boldres{\scalebox{0.78}{0.188}} &\secondres{\scalebox{0.78}{0.084}} &\secondres{\scalebox{0.78}{0.192}} &{\scalebox{0.78}{0.095}} &{\scalebox{0.78}{0.205}} &\scalebox{0.78}{0.258} &\scalebox{0.78}{0.348} &\scalebox{0.78}{0.153} &\scalebox{0.78}{0.275} &\scalebox{0.78}{0.131} &\scalebox{0.78}{0.256} & \scalebox{0.78}{0.292}& \scalebox{0.78}{0.398} &\scalebox{0.78}{0.103} &\scalebox{0.78}{0.215} &\scalebox{0.78}{0.224} &\scalebox{0.78}{0.340} &\secondres{\scalebox{0.78}{0.084}} &{\scalebox{0.78}{0.193}} &\scalebox{0.78}{0.177} &\scalebox{0.78}{0.293} &\scalebox{0.78}{0.104} &\scalebox{0.78}{0.216} \\% &\scalebox{0.78}{0.117} &\scalebox{0.78}{0.227}\\
     
					& \scalebox{0.78}{48} & \boldres{\scalebox{0.78}{0.099}} &\boldres{\scalebox{0.78}{0.210}} &\secondres{\scalebox{0.78}{0.101}} &{\scalebox{0.78}{0.212}} &{\scalebox{0.78}{0.120}} &{\scalebox{0.78}{0.233}} &\scalebox{0.78}{0.572} &\scalebox{0.78}{0.544} &\scalebox{0.78}{0.229} &\scalebox{0.78}{0.339} &\scalebox{0.78}{0.205} &\scalebox{0.78}{0.326} & \scalebox{0.78}{0.409}& \scalebox{0.78}{0.478} &\scalebox{0.78}{0.136} &\scalebox{0.78}{0.250} &\scalebox{0.78}{0.355} &\scalebox{0.78}{0.437} &\boldres{\scalebox{0.78}{0.099}} &\secondres{\scalebox{0.78}{0.211}} &\scalebox{0.78}{0.270} &\scalebox{0.78}{0.368} &\scalebox{0.78}{0.137} &\scalebox{0.78}{0.251}  \\% &\scalebox{0.78}{0.126} &\scalebox{0.78}{0.239}\\
     
					& \scalebox{0.78}{96} & \secondres{\scalebox{0.78}{0.125}} &{\scalebox{0.78}{0.240}} &{\scalebox{0.78}{0.127}} &\secondres{\scalebox{0.78}{0.236}} &{\scalebox{0.78}{0.150}} &{\scalebox{0.78}{0.262}} &\scalebox{0.78}{1.137} &\scalebox{0.78}{0.820} &\scalebox{0.78}{0.291} &\scalebox{0.78}{0.389} &\scalebox{0.78}{0.402} &\scalebox{0.78}{0.457} & \scalebox{0.78}{0.492}& \scalebox{0.78}{0.532} &\scalebox{0.78}{0.190} &\scalebox{0.78}{0.303} &\scalebox{0.78}{0.452} &\scalebox{0.78}{0.504} &\boldres{\scalebox{0.78}{0.114}} &\boldres{\scalebox{0.78}{0.227}} &\scalebox{0.78}{0.341} &\scalebox{0.78}{0.427} &\scalebox{0.78}{0.186} &\scalebox{0.78}{0.297}  \\% &\scalebox{0.78}{0.128} &\scalebox{0.78}{0.242}\\
     
					\cmidrule(lr){2-24}
					& \scalebox{0.78}{Avg} & \secondres{\scalebox{0.78}{0.094}} &\secondres{\scalebox{0.78}{0.203}} &{\scalebox{0.78}{0.096}} &{\scalebox{0.78}{0.204}} &{\scalebox{0.78}{0.111}} &{\scalebox{0.78}{0.221}} &\scalebox{0.78}{0.526} &\scalebox{0.78}{0.491} &\scalebox{0.78}{0.195} &\scalebox{0.78}{0.307} &\scalebox{0.78}{0.209} &\scalebox{0.78}{0.314} & \scalebox{0.78}{0.353}& \scalebox{0.78}{0.437} &\scalebox{0.78}{0.129} &\scalebox{0.78}{0.241} &\scalebox{0.78}{0.295} &\scalebox{0.78}{0.388} &\boldres{\scalebox{0.78}{0.092}} &\boldres{\scalebox{0.78}{0.202}} &\scalebox{0.78}{0.231} &\scalebox{0.78}{0.337} &\scalebox{0.78}{0.127} &\scalebox{0.78}{0.240} \\% &\scalebox{0.78}{0.121} &\scalebox{0.78}{0.232}\\
					
					\midrule
					\multirow{5}{*}{\update{\rotatebox{90}{\scalebox{0.95}{PEMS07}}}}
					&  \scalebox{0.78}{12} & \boldres{\scalebox{0.78}{0.057}} &\boldres{\scalebox{0.78}{0.153}} &\secondres{\scalebox{0.78}{0.060}} &\secondres{\scalebox{0.78}{0.157}} &{\scalebox{0.78}{0.067}} &{\scalebox{0.78}{0.165}} &\scalebox{0.78}{0.118} &\scalebox{0.78}{0.235} &\scalebox{0.78}{0.095} &\scalebox{0.78}{0.207} &\scalebox{0.78}{0.094} &\scalebox{0.78}{0.200} & \scalebox{0.78}{0.173}& \scalebox{0.78}{0.304} &\scalebox{0.78}{0.082} &\scalebox{0.78}{0.181} &\scalebox{0.78}{0.115} &\scalebox{0.78}{0.242} &{\scalebox{0.78}{0.068}} &{\scalebox{0.78}{0.171}} &\scalebox{0.78}{0.109} &\scalebox{0.78}{0.225} &\scalebox{0.78}{0.083} &\scalebox{0.78}{0.185}  \\% &\scalebox{0.78}{0.173} &\scalebox{0.78}{0.243} \\
     
					& \scalebox{0.78}{24} & \boldres{\scalebox{0.78}{0.069}} &\boldres{\scalebox{0.78}{0.165}} &\secondres{\scalebox{0.78}{0.077}} &\secondres{\scalebox{0.78}{0.178}} &{\scalebox{0.78}{0.088}} &{\scalebox{0.78}{0.190}} &\scalebox{0.78}{0.242} &\scalebox{0.78}{0.341} &\scalebox{0.78}{0.150} &\scalebox{0.78}{0.262} &\scalebox{0.78}{0.139} &\scalebox{0.78}{0.247} & \scalebox{0.78}{0.271}& \scalebox{0.78}{0.383} &\scalebox{0.78}{0.101} &\scalebox{0.78}{0.204} &\scalebox{0.78}{0.210} &\scalebox{0.78}{0.329} &{\scalebox{0.78}{0.119}} &{\scalebox{0.78}{0.225}} &\scalebox{0.78}{0.125} &\scalebox{0.78}{0.244} &\scalebox{0.78}{0.102} &\scalebox{0.78}{0.207} \\% &\scalebox{0.78}{0.178} &\scalebox{0.78}{0.247}\\
     
					& \scalebox{0.78}{48} & \boldres{\scalebox{0.78}{0.084}} &\boldres{\scalebox{0.78}{0.183}} &\secondres{\scalebox{0.78}{0.095}} &\secondres{\scalebox{0.78}{0.197}} &{\scalebox{0.78}{0.110}} &{\scalebox{0.78}{0.215}} &\scalebox{0.78}{0.562} &\scalebox{0.78}{0.541} &\scalebox{0.78}{0.253} &\scalebox{0.78}{0.340} &\scalebox{0.78}{0.311} &\scalebox{0.78}{0.369} & \scalebox{0.78}{0.446}& \scalebox{0.78}{0.495} &\scalebox{0.78}{0.134} &\scalebox{0.78}{0.238} &\scalebox{0.78}{0.398} &\scalebox{0.78}{0.458} &{\scalebox{0.78}{0.149}} &{\scalebox{0.78}{0.237}} &\scalebox{0.78}{0.165} &\scalebox{0.78}{0.288} &\scalebox{0.78}{0.136} &\scalebox{0.78}{0.240} \\% &\scalebox{0.78}{0.185} &\scalebox{0.78}{0.256}\\
     
					& \scalebox{0.78}{96} & \boldres{\scalebox{0.78}{0.105}} &\boldres{\scalebox{0.78}{0.204}} &\secondres{\scalebox{0.78}{0.118}} &\secondres{\scalebox{0.78}{0.218}} &{\scalebox{0.78}{0.139}} &{\scalebox{0.78}{0.245}} &\scalebox{0.78}{1.096} &\scalebox{0.78}{0.795} &\scalebox{0.78}{0.346} &\scalebox{0.78}{0.404} &\scalebox{0.78}{0.396} &\scalebox{0.78}{0.442} & \scalebox{0.78}{0.628}& \scalebox{0.78}{0.577} &\scalebox{0.78}{0.181} &\scalebox{0.78}{0.279} &\scalebox{0.78}{0.594} &\scalebox{0.78}{0.553} &{\scalebox{0.78}{0.141}} &{\scalebox{0.78}{0.234}} &\scalebox{0.78}{0.262} &\scalebox{0.78}{0.376} &\scalebox{0.78}{0.187} &\scalebox{0.78}{0.287} \\% &\scalebox{0.78}{0.195} &\scalebox{0.78}{0.269}\\
     
					\cmidrule(lr){2-24}
					& \scalebox{0.78}{Avg} & \boldres{\scalebox{0.78}{0.079}} &\boldres{\scalebox{0.78}{0.176}} &\secondres{\scalebox{0.78}{0.088}} &\secondres{\scalebox{0.78}{0.188}} &{\scalebox{0.78}{0.101}} &{\scalebox{0.78}{0.204}} &\scalebox{0.78}{0.504} &\scalebox{0.78}{0.478} &\scalebox{0.78}{0.211} &\scalebox{0.78}{0.303} &\scalebox{0.78}{0.235} &\scalebox{0.78}{0.315} & \scalebox{0.78}{0.380}& \scalebox{0.78}{0.440} &\scalebox{0.78}{0.124} &\scalebox{0.78}{0.225} &\scalebox{0.78}{0.329} &\scalebox{0.78}{0.395} &{\scalebox{0.78}{0.119}} &{\scalebox{0.78}{0.234}} &\scalebox{0.78}{0.165} &\scalebox{0.78}{0.283} &\scalebox{0.78}{0.127} &\scalebox{0.78}{0.230} \\% &\scalebox{0.78}{0.183} &\scalebox{0.78}{0.254}\\
					
					\midrule
					\multirow{5}{*}{\update{\rotatebox{90}{\scalebox{0.95}{PEMS08}}}}
					&  \scalebox{0.78}{12} & \boldres{\scalebox{0.78}{0.072}} &\boldres{\scalebox{0.78}{0.171}} &\secondres{\scalebox{0.78}{0.076}} &\secondres{\scalebox{0.78}{0.178}} &{\scalebox{0.78}{0.079}} &{\scalebox{0.78}{0.182}} &\scalebox{0.78}{0.133} &\scalebox{0.78}{0.247} &\scalebox{0.78}{0.168} &\scalebox{0.78}{0.232} &\scalebox{0.78}{0.165} &\scalebox{0.78}{0.214} & \scalebox{0.78}{0.227}& \scalebox{0.78}{0.343} &\scalebox{0.78}{0.112} &\scalebox{0.78}{0.212} &\scalebox{0.78}{0.154} &\scalebox{0.78}{0.276} &{\scalebox{0.78}{0.087}} &{\scalebox{0.78}{0.184}} &\scalebox{0.78}{0.173} &\scalebox{0.78}{0.273} &\scalebox{0.78}{0.109} &\scalebox{0.78}{0.207} \\% &\scalebox{0.78}{0.296} &\scalebox{0.78}{0.312}\\
     
					& \scalebox{0.78}{24} & \boldres{\scalebox{0.78}{0.091}} &\boldres{\scalebox{0.78}{0.192}} &\secondres{\scalebox{0.78}{0.110}} &\secondres{\scalebox{0.78}{0.216}} &{\scalebox{0.78}{0.115}} &{\scalebox{0.78}{0.219}} &\scalebox{0.78}{0.249} &\scalebox{0.78}{0.343} &\scalebox{0.78}{0.224} &\scalebox{0.78}{0.281} &\scalebox{0.78}{0.215} &\scalebox{0.78}{0.260} & \scalebox{0.78}{0.318}& \scalebox{0.78}{0.409} &\scalebox{0.78}{0.141} &\scalebox{0.78}{0.238} &\scalebox{0.78}{0.248} &\scalebox{0.78}{0.353} &{\scalebox{0.78}{0.122}} &{\scalebox{0.78}{0.221}} &\scalebox{0.78}{0.210} &\scalebox{0.78}{0.301} &\scalebox{0.78}{0.140} &\scalebox{0.78}{0.236} \\% &\scalebox{0.78}{0.327} &\scalebox{0.78}{0.318}\\
     
					& \scalebox{0.78}{48} & \boldres{\scalebox{0.78}{0.121}} &\boldres{\scalebox{0.78}{0.221}} &\secondres{\scalebox{0.78}{0.165}} &{\scalebox{0.78}{0.252}} &{\scalebox{0.78}{0.186}} &\secondres{\scalebox{0.78}{0.235}} &\scalebox{0.78}{0.569} &\scalebox{0.78}{0.544} &\scalebox{0.78}{0.321} &\scalebox{0.78}{0.354} &\scalebox{0.78}{0.315} &\scalebox{0.78}{0.355} & \scalebox{0.78}{0.497}& \scalebox{0.78}{0.510} &\scalebox{0.78}{0.198} &\scalebox{0.78}{0.283} &\scalebox{0.78}{0.440} &\scalebox{0.78}{0.470} &{\scalebox{0.78}{0.189}} &x{\scalebox{0.78}{0.270}} &\scalebox{0.78}{0.320} &\scalebox{0.78}{0.394} &\scalebox{0.78}{0.211} &\scalebox{0.78}{0.294} \\% &\scalebox{0.78}{0.387} &\scalebox{0.78}{0.365}\\
     
					& \scalebox{0.78}{96} & \boldres{\scalebox{0.78}{0.175}} &\boldres{\scalebox{0.78}{0.261}} &{\scalebox{0.78}{0.274}} &{\scalebox{0.78}{0.327}} &\secondres{\scalebox{0.78}{0.221}} &\secondres{\scalebox{0.78}{0.267}} &\scalebox{0.78}{1.166} &\scalebox{0.78}{0.814} &\scalebox{0.78}{0.408} &\scalebox{0.78}{0.417} &\scalebox{0.78}{0.377} &\scalebox{0.78}{0.397} & \scalebox{0.78}{0.721}& \scalebox{0.78}{0.592} &\scalebox{0.78}{0.320} &\scalebox{0.78}{0.351} &\scalebox{0.78}{0.674} &\scalebox{0.78}{0.565} &{\scalebox{0.78}{0.236}} &{\scalebox{0.78}{0.300}} &\scalebox{0.78}{0.442} &\scalebox{0.78}{0.465} &\scalebox{0.78}{0.345} &\scalebox{0.78}{0.367}  \\% &\scalebox{0.78}{0.455} &\scalebox{0.78}{0.407}\\
     
					\cmidrule(lr){2-24}
					& \scalebox{0.78}{Avg} & \boldres{\scalebox{0.78}{0.115}} &\boldres{\scalebox{0.78}{0.211}} &{\scalebox{0.78}{0.156}} &{\scalebox{0.78}{0.243}} &\secondres{\scalebox{0.78}{0.150}} &\secondres{\scalebox{0.78}{0.226}} &\scalebox{0.78}{0.529} &\scalebox{0.78}{0.487} &\scalebox{0.78}{0.280} &\scalebox{0.78}{0.321} &\scalebox{0.78}{0.268} &\scalebox{0.78}{0.307} & \scalebox{0.78}{0.441}& \scalebox{0.78}{0.464} &\scalebox{0.78}{0.193} &\scalebox{0.78}{0.271} &\scalebox{0.78}{0.379} &\scalebox{0.78}{0.416} &{\scalebox{0.78}{0.158}} &{\scalebox{0.78}{0.244}} &\scalebox{0.78}{0.286} &\scalebox{0.78}{0.358} &\scalebox{0.78}{0.201} &\scalebox{0.78}{0.276}  \\% &\scalebox{0.78}{0.366} &\scalebox{0.78}{0.350}\\
     
					\midrule

                    \multirow{5}{*}{\rotatebox{90}{\scalebox{0.95}{Electricity}}} 
					&  \scalebox{0.78}{96} & \boldres{\scalebox{0.78}{0.137}} &\secondres{\scalebox{0.78}{0.238}} &\secondres{\scalebox{0.78}{0.139}} &\boldres{\scalebox{0.78}{0.235}} & {\scalebox{0.78}{0.148}} & {\scalebox{0.78}{0.240}} & \scalebox{0.78}{0.201} & \scalebox{0.78}{0.281} & \scalebox{0.78}{0.195} & \scalebox{0.78}{0.285} & \scalebox{0.78}{0.219} & \scalebox{0.78}{0.314} & \scalebox{0.78}{0.237} & \scalebox{0.78}{0.329} &{\scalebox{0.78}{0.168}} &{\scalebox{0.78}{0.272}} &\scalebox{0.78}{0.197} &\scalebox{0.78}{0.282} & \scalebox{0.78}{0.247} & \scalebox{0.78}{0.345} &\scalebox{0.78}{0.193} &\scalebox{0.78}{0.308} &{\scalebox{0.78}{0.169}} &{\scalebox{0.78}{0.273}}  \\ %&\scalebox{0.78}{0.274} &\scalebox{0.78}{0.368} \\
     
					& \scalebox{0.78}{192}  & \boldres{\scalebox{0.78}{0.157}} &\secondres{\scalebox{0.78}{0.255}} &\secondres{\scalebox{0.78}{0.161}} &{\scalebox{0.78}{0.258}} & {\scalebox{0.78}{0.162}} & \boldres{\scalebox{0.78}{0.253}} & \scalebox{0.78}{0.201} & {\scalebox{0.78}{0.283}} & \scalebox{0.78}{0.199} & \scalebox{0.78}{0.289} & \scalebox{0.78}{0.231} & \scalebox{0.78}{0.322} & \scalebox{0.78}{0.236} & \scalebox{0.78}{0.330} &{\scalebox{0.78}{0.184}} &\scalebox{0.78}{0.289} &\scalebox{0.78}{0.196} &{\scalebox{0.78}{0.285}} & \scalebox{0.78}{0.257} & \scalebox{0.78}{0.355} &\scalebox{0.78}{0.201} &\scalebox{0.78}{0.315} &{\scalebox{0.78}{0.182}} &\scalebox{0.78}{0.286} \\ %&\scalebox{0.78}{0.296} &\scalebox{0.78}{0.386} \\
     
					& \scalebox{0.78}{336}  & \boldres{\scalebox{0.78}{0.174}} &\secondres{\scalebox{0.78}{0.276}} &{\scalebox{0.78}{0.181}} &{\scalebox{0.78}{0.278}} & \secondres{\scalebox{0.78}{0.178}} & \boldres{\scalebox{0.78}{0.269}} & \scalebox{0.78}{0.215} & {\scalebox{0.78}{0.298}} & \scalebox{0.78}{0.215} & \scalebox{0.78}{0.305} & \scalebox{0.78}{0.246} & \scalebox{0.78}{0.337} & \scalebox{0.78}{0.249} & \scalebox{0.78}{0.344} &{\scalebox{0.78}{0.198}} &{\scalebox{0.78}{0.300}} &\scalebox{0.78}{0.209} &{\scalebox{0.78}{0.301}} & \scalebox{0.78}{0.269} & \scalebox{0.78}{0.369} &\scalebox{0.78}{0.214} &\scalebox{0.78}{0.329} &{\scalebox{0.78}{0.200}} &\scalebox{0.78}{0.304} \\ %&\scalebox{0.78}{0.300} &\scalebox{0.78}{0.394} \\
     
					& \scalebox{0.78}{720}  & \secondres{\scalebox{0.78}{0.208}} &\secondres{\scalebox{0.78}{0.308}} &\boldres{\scalebox{0.78}{0.201}} &\boldres{\scalebox{0.78}{0.298}} & {\scalebox{0.78}{0.225}} & {\scalebox{0.78}{0.317}} & \scalebox{0.78}{0.257} & \scalebox{0.78}{0.331} & \scalebox{0.78}{0.256} & \scalebox{0.78}{0.337} & \scalebox{0.78}{0.280} & \scalebox{0.78}{0.363} & \scalebox{0.78}{0.284} & \scalebox{0.78}{0.373} &{\scalebox{0.78}{0.220}} &{\scalebox{0.78}{0.320}} &\scalebox{0.78}{0.245} &\scalebox{0.78}{0.333} & \scalebox{0.78}{0.299} & \scalebox{0.78}{0.390} &\scalebox{0.78}{0.246} &\scalebox{0.78}{0.355} &{\scalebox{0.78}{0.222}} &{\scalebox{0.78}{0.321}}  \\ %&\scalebox{0.78}{0.373} &\scalebox{0.78}{0.439} \\
     
					\cmidrule(lr){2-24}
					& \scalebox{0.78}{Avg} & \boldres{\scalebox{0.78}{0.169}} &\secondres{\scalebox{0.78}{0.269}} &\secondres{\scalebox{0.78}{0.171}} &\boldres{\scalebox{0.78}{0.267}} & {\scalebox{0.78}{0.178}} & {\scalebox{0.78}{0.270}} & \scalebox{0.78}{0.219} & \scalebox{0.78}{0.298} & \scalebox{0.78}{0.216} & \scalebox{0.78}{0.304} & \scalebox{0.78}{0.244} & \scalebox{0.78}{0.334} & \scalebox{0.78}{0.251} & \scalebox{0.78}{0.344} &{\scalebox{0.78}{0.192}} &{\scalebox{0.78}{0.295}} &\scalebox{0.78}{0.212} &\scalebox{0.78}{0.300} & \scalebox{0.78}{0.268} & \scalebox{0.78}{0.365} &\scalebox{0.78}{0.214} &\scalebox{0.78}{0.327} &{\scalebox{0.78}{0.193}} &{\scalebox{0.78}{0.296}} \\ %&\scalebox{0.78}{0.311} &\scalebox{0.78}{0.397} \\
					\midrule

					\multirow{5}{*}{\rotatebox{90}{\scalebox{0.95}{SML2010}}}
					&  \scalebox{0.78}{48} & \boldres{\scalebox{0.78}{0.244}} &\boldres{\scalebox{0.78}{0.279}} &\secondres{\scalebox{0.78}{0.266}} &\secondres{\scalebox{0.78}{0.290}} & {\scalebox{0.78}{0.293}} & {\scalebox{0.78}{0.300}} & {\scalebox{0.78}{0.289}} & \scalebox{0.78}{0.327} & \scalebox{0.78}{0.295} & \scalebox{0.78}{0.323} & {\scalebox{0.78}{0.277}} & \scalebox{0.78}{0.332} & \scalebox{0.78}{0.408} & \scalebox{0.78}{0.411} &{\scalebox{0.78}{0.313}} &{\scalebox{0.78}{0.307}} &\scalebox{0.78}{0.324} &\scalebox{0.78}{0.363} & \scalebox{0.78}{N/A} & \scalebox{0.78}{N/A} &\scalebox{0.78}{0.313} &\scalebox{0.78}{0.370} &{\scalebox{0.78}{0.399}} &{\scalebox{0.78}{0.376}}  \\  
     
					&  \scalebox{0.78}{96} & \boldres{\scalebox{0.78}{0.319}} &\boldres{\scalebox{0.78}{0.338}} &{\scalebox{0.78}{0.415}} &{\scalebox{0.78}{0.364}} & {\scalebox{0.78}{0.346}} & {\scalebox{0.78}{0.347}} & {\scalebox{0.78}{0.327}} & \scalebox{0.78}{0.357} & \scalebox{0.78}{0.345} & \secondres{\scalebox{0.78}{0.342}} & \secondres{\scalebox{0.78}{0.323}} & \scalebox{0.78}{0.365} & \scalebox{0.78}{0.436} & \scalebox{0.78}{0.426} &{\scalebox{0.78}{0.382}} &{\scalebox{0.78}{0.367}} &\scalebox{0.78}{0.359} &\scalebox{0.78}{0.387} & \scalebox{0.78}{N/A} & \scalebox{0.78}{N/A} &\scalebox{0.78}{0.374} &\scalebox{0.78}{0.411} &{\scalebox{0.78}{0.454}} &{\scalebox{0.78}{0.411}}  \\ 
     
					&  \scalebox{0.78}{192} & \boldres{\scalebox{0.78}{0.386}} &\boldres{\scalebox{0.78}{0.384}} &{\scalebox{0.78}{0.457}} &{\scalebox{0.78}{0.411}} & {\scalebox{0.78}{0.447}} & {\scalebox{0.78}{0.413}} & {\scalebox{0.78}{0.403}} & \scalebox{0.78}{0.404} & \secondres{\scalebox{0.78}{0.401}} & \secondres{\scalebox{0.78}{0.396}} & \scalebox{0.78}{0.483} & \scalebox{0.78}{0.466} & \scalebox{0.78}{0.499} & \scalebox{0.78}{0.461} &{\scalebox{0.78}{0.471}} &{\scalebox{0.78}{0.417}} &\scalebox{0.78}{0.428} &\scalebox{0.78}{0.429} & \scalebox{0.78}{N/A} & \scalebox{0.78}{N/A} &\scalebox{0.78}{0.471} &\scalebox{0.78}{0.463} &{\scalebox{0.78}{0.550}} &{\scalebox{0.78}{0.455}} \\
     
					&  \scalebox{0.78}{336} & \boldres{\scalebox{0.78}{0.418}} &\boldres{\scalebox{0.78}{0.411}} &{\scalebox{0.78}{0.523}} &{\scalebox{0.78}{0.451}} & {\scalebox{0.78}{0.576}} & {\scalebox{0.78}{0.485}} & \secondres{\scalebox{0.78}{0.440}} & \secondres{\scalebox{0.78}{0.432}} & \scalebox{0.78}{0.480} & \scalebox{0.78}{0.441} & \scalebox{0.78}{0.823} & \scalebox{0.78}{0.668} & \scalebox{0.78}{0.568} & \scalebox{0.78}{0.510} &{\scalebox{0.78}{0.596}} &{\scalebox{0.78}{0.486}} &\scalebox{0.78}{0.482} &\scalebox{0.78}{0.464} & \scalebox{0.78}{N/A} & \scalebox{0.78}{N/A} &\scalebox{0.78}{0.905} &\scalebox{0.78}{0.678} &{\scalebox{0.78}{0.798}} &{\scalebox{0.78}{0.551}} \\
					
					\cmidrule(lr){2-24}
					& \scalebox{0.78}{Avg} & \boldres{\scalebox{0.78}{0.342}} &\boldres{\scalebox{0.78}{0.353}} &{\scalebox{0.78}{0.415}} &{\scalebox{0.78}{0.379}} & {\scalebox{0.78}{0.416}} & {\scalebox{0.78}{0.386}} & \secondres{\scalebox{0.78}{0.365}} & \scalebox{0.78}{0.380} & \scalebox{0.78}{0.380} & \secondres{\scalebox{0.78}{0.376}} & \scalebox{0.78}{0.477} & \scalebox{0.78}{0.458} & \scalebox{0.78}{0.478} & \scalebox{0.78}{0.452} &{\scalebox{0.78}{0.441}} &{\scalebox{0.78}{0.394}} &\scalebox{0.78}{0.398} &\scalebox{0.78}{0.411} & \scalebox{0.78}{N/A} & \scalebox{0.78}{N/A} &\scalebox{0.78}{0.516} &\scalebox{0.78}{0.481} &{\scalebox{0.78}{0.550}} &{\scalebox{0.78}{0.448}} \\ %&\scalebox{0.78}{0.311} &\scalebox{0.78}{0.397} \\
					\midrule
					
					\multirow{5}{*}{\rotatebox{90}{\scalebox{0.95}{Weather}}} 
					&  \scalebox{0.78}{96}  & \secondres{\scalebox{0.78}{0.164}} &{\scalebox{0.78}{0.222}} &{\scalebox{0.78}{0.165}} &\boldres{\scalebox{0.78}{0.209}} & \scalebox{0.78}{0.174} & \secondres{\scalebox{0.78}{0.214}} & \scalebox{0.78}{0.192} & \scalebox{0.78}{0.232} & \scalebox{0.78}{0.177} & {\scalebox{0.78}{0.218}} & \boldres{\scalebox{0.78}{0.158}} & \scalebox{0.78}{0.230}  & \scalebox{0.78}{0.202} & \scalebox{0.78}{0.261} &{\scalebox{0.78}{0.172}} &{\scalebox{0.78}{0.220}} & \scalebox{0.78}{0.196} &\scalebox{0.78}{0.255} & \scalebox{0.78}{0.221} & \scalebox{0.78}{0.306} & \scalebox{0.78}{0.217} &\scalebox{0.78}{0.296} & {\scalebox{0.78}{0.173}} &{\scalebox{0.78}{0.223}}  \\ %& \scalebox{0.78}{0.300} &\scalebox{0.78}{0.384}  \\
     
					& \scalebox{0.78}{192}  & \secondres{\scalebox{0.78}{0.214}} &{\scalebox{0.78}{0.265}} &{\scalebox{0.78}{0.215}} &\secondres{\scalebox{0.78}{0.255}} & \scalebox{0.78}{0.221} & \boldres{\scalebox{0.78}{0.254}} & \scalebox{0.78}{0.240} & \scalebox{0.78}{0.271} & \scalebox{0.78}{0.225} & {\scalebox{0.78}{0.259}} & \boldres{\scalebox{0.78}{0.206}} & \scalebox{0.78}{0.277} & \scalebox{0.78}{0.242} & \scalebox{0.78}{0.298} &{\scalebox{0.78}{0.219}} &{\scalebox{0.78}{0.261}}  & \scalebox{0.78}{0.237} &\scalebox{0.78}{0.296} & \scalebox{0.78}{0.261} & \scalebox{0.78}{0.340} & \scalebox{0.78}{0.276} &\scalebox{0.78}{0.336} & \scalebox{0.78}{0.245} &\scalebox{0.78}{0.285}  \\ %& \scalebox{0.78}{0.598} &\scalebox{0.78}{0.544} \\
     
					& \scalebox{0.78}{336}  & \boldres{\scalebox{0.78}{0.267}} &{\scalebox{0.78}{0.319}} &{\scalebox{0.78}{0.273}} &\boldres{\scalebox{0.78}{0.296}} & {\scalebox{0.78}{0.278}} & \boldres{\scalebox{0.78}{0.296}} & \scalebox{0.78}{0.292} & \scalebox{0.78}{0.307} & \scalebox{0.78}{0.278} & \secondres{\scalebox{0.78}{0.297}} & \secondres{\scalebox{0.78}{0.272}} & \scalebox{0.78}{0.335} & \scalebox{0.78}{0.287} & \scalebox{0.78}{0.335} &{\scalebox{0.78}{0.280}} &{\scalebox{0.78}{0.306}} & \scalebox{0.78}{0.283} &\scalebox{0.78}{0.335} & \scalebox{0.78}{0.309} & \scalebox{0.78}{0.378} & \scalebox{0.78}{0.339} &\scalebox{0.78}{0.380} & \scalebox{0.78}{0.321} &\scalebox{0.78}{0.338} \\ % &\scalebox{0.78}{0.578} &\scalebox{0.78}{0.523} \\
     
					& \scalebox{0.78}{720}  & \boldres{\scalebox{0.78}{0.341}} &{\scalebox{0.78}{0.364}} &{\scalebox{0.78}{0.353}} &{\scalebox{0.78}{0.349}} & \scalebox{0.78}{0.358} & \boldres{\scalebox{0.78}{0.347}} & \scalebox{0.78}{0.364} & \scalebox{0.78}{0.353} & {\scalebox{0.78}{0.354}} & \secondres{\scalebox{0.78}{0.348}} & \scalebox{0.78}{0.398} & \scalebox{0.78}{0.418} & \secondres{\scalebox{0.78}{0.351}} & \scalebox{0.78}{0.386} &\scalebox{0.78}{0.365} &{\scalebox{0.78}{0.359}} & {\scalebox{0.78}{0.345}} &{\scalebox{0.78}{0.381}} & \scalebox{0.78}{0.377} & \scalebox{0.78}{0.427} & \scalebox{0.78}{0.403} &\scalebox{0.78}{0.428} & \scalebox{0.78}{0.414} &\scalebox{0.78}{0.410}  \\ %& \scalebox{0.78}{1.059} &\scalebox{0.78}{0.741} \\
     
					\cmidrule(lr){2-24}
					& \scalebox{0.78}{Avg}  & \boldres{\scalebox{0.78}{0.247}} &{\scalebox{0.78}{0.293}} &\secondres{\scalebox{0.78}{0.252}} &\boldres{\scalebox{0.78}{0.277}} & {\scalebox{0.78}{0.258}} & \secondres{\scalebox{0.78}{0.279}} & \scalebox{0.78}{0.272} & \scalebox{0.78}{0.291} & {\scalebox{0.78}{0.259}} & {\scalebox{0.78}{0.281}} & \scalebox{0.78}{0.259} & \scalebox{0.78}{0.315} & \scalebox{0.78}{0.271} & \scalebox{0.78}{0.320} &{\scalebox{0.78}{0.259}} &{\scalebox{0.78}{0.287}} &\scalebox{0.78}{0.265} &\scalebox{0.78}{0.317} & \scalebox{0.78}{0.292} & \scalebox{0.78}{0.363} &\scalebox{0.78}{0.309} &\scalebox{0.78}{0.360} &\scalebox{0.78}{0.288} &\scalebox{0.78}{0.314} \\ %&\scalebox{0.78}{0.634} &\scalebox{0.78}{0.548} \\
					\midrule
					
					\multirow{5}{*}{\rotatebox{90}{\scalebox{0.95}{Solar-Energy}}} 
					&  \scalebox{0.78}{96}  & \boldres{\scalebox{0.78}{0.195}} &{\scalebox{0.78}{0.251}} &{\scalebox{0.78}{0.208}} &\secondres{\scalebox{0.78}{0.246}} &\secondres{\scalebox{0.78}{0.203}} &\boldres{\scalebox{0.78}{0.237}} & \scalebox{0.78}{0.322} & \scalebox{0.78}{0.339} & {\scalebox{0.78}{0.234}} & {\scalebox{0.78}{0.286}} &\scalebox{0.78}{0.310} &\scalebox{0.78}{0.331} &\scalebox{0.78}{0.312} &\scalebox{0.78}{0.399} &\scalebox{0.78}{0.250} &\scalebox{0.78}{0.292} &\scalebox{0.78}{0.290} &\scalebox{0.78}{0.378} &\scalebox{0.78}{0.237} &\scalebox{0.78}{0.344} &\scalebox{0.78}{0.242} &\scalebox{0.78}{0.342} &\scalebox{0.78}{0.215} &\scalebox{0.78}{0.249} \\ % &\scalebox{0.78}{0.236} &\secondres{\scalebox{0.78}{0.259}} \\
     
					& \scalebox{0.78}{192} & \boldres{\scalebox{0.78}{0.219}} &{\scalebox{0.78}{0.275}} &{\scalebox{0.78}{0.240}} &\secondres{\scalebox{0.78}{0.272}} &\secondres{\scalebox{0.78}{0.233}} &\boldres{\scalebox{0.78}{0.261}} & \scalebox{0.78}{0.359} & \scalebox{0.78}{0.356}& {\scalebox{0.78}{0.267}} & {\scalebox{0.78}{0.310}} &\scalebox{0.78}{0.734} &\scalebox{0.78}{0.725} &\scalebox{0.78}{0.339} &\scalebox{0.78}{0.416} &\scalebox{0.78}{0.296} &\scalebox{0.78}{0.318} &\scalebox{0.78}{0.320} &\scalebox{0.78}{0.398} &\scalebox{0.78}{0.280} &\scalebox{0.78}{0.380} &\scalebox{0.78}{0.285} &\scalebox{0.78}{0.380} &\scalebox{0.78}{0.254} &\scalebox{0.78}{0.272} \\ %&\boldres{\scalebox{0.78}{0.217}} &\secondres{\scalebox{0.78}{0.269}} \\
     
					& \scalebox{0.78}{336}  & \boldres{\scalebox{0.78}{0.218}} &\secondres{\scalebox{0.78}{0.276}} &{\scalebox{0.78}{0.262}} &{\scalebox{0.78}{0.290}} &\secondres{\scalebox{0.78}{0.248}} &\boldres{\scalebox{0.78}{0.273}} & \scalebox{0.78}{0.397} & \scalebox{0.78}{0.369}& {\scalebox{0.78}{0.290}}  &{\scalebox{0.78}{0.315}} &\scalebox{0.78}{0.750} &\scalebox{0.78}{0.735} &\scalebox{0.78}{0.368} &\scalebox{0.78}{0.430} &\scalebox{0.78}{0.319} &\scalebox{0.78}{0.330} &\scalebox{0.78}{0.353} &\scalebox{0.78}{0.415} &\scalebox{0.78}{0.304} &\scalebox{0.78}{0.389} &\scalebox{0.78}{0.282} &\scalebox{0.78}{0.376} &\scalebox{0.78}{0.290} &\scalebox{0.78}{0.296} \\ %&\secondres{\scalebox{0.78}{0.249}} &\secondres{\scalebox{0.78}{0.283}}\\
     
					& \scalebox{0.78}{720}  & \boldres{\scalebox{0.78}{0.221}} &\secondres{\scalebox{0.78}{0.279}} &{\scalebox{0.78}{0.267}} &{\scalebox{0.78}{0.293}} &\secondres{\scalebox{0.78}{0.249}} &\boldres{\scalebox{0.78}{0.275}} & \scalebox{0.78}{0.397} & \scalebox{0.78}{0.356} & {\scalebox{0.78}{0.289}} &{\scalebox{0.78}{0.317}} &\scalebox{0.78}{0.769} &\scalebox{0.78}{0.765} &\scalebox{0.78}{0.370} &\scalebox{0.78}{0.425} &\scalebox{0.78}{0.338} &\scalebox{0.78}{0.337} &\scalebox{0.78}{0.356} &\scalebox{0.78}{0.413} &\scalebox{0.78}{0.308} &\scalebox{0.78}{0.388} &\scalebox{0.78}{0.357} &\scalebox{0.78}{0.427} &\scalebox{0.78}{0.285} &\scalebox{0.78}{0.295}  \\ %&\boldres{\scalebox{0.78}{0.241}} &\boldres{\scalebox{0.78}{0.317}}\\
     
					\cmidrule(lr){2-24}
					& \scalebox{0.78}{Avg}  & \boldres{\scalebox{0.78}{0.213}} &\secondres{\scalebox{0.78}{0.270}} &{\scalebox{0.78}{0.244}} &{\scalebox{0.78}{0.275}} &\secondres{\scalebox{0.78}{0.233}} &\boldres{\scalebox{0.78}{0.262}} & \scalebox{0.78}{0.369} & \scalebox{0.78}{0.356} &{\scalebox{0.78}{0.270}} &{\scalebox{0.78}{0.307}} &\scalebox{0.78}{0.641} &\scalebox{0.78}{0.639} &\scalebox{0.78}{0.347} &\scalebox{0.78}{0.417} &\scalebox{0.78}{0.301} &\scalebox{0.78}{0.319} &\scalebox{0.78}{0.330} &\scalebox{0.78}{0.401} &\scalebox{0.78}{0.282} &\scalebox{0.78}{0.375} &\scalebox{0.78}{0.291} &\scalebox{0.78}{0.381} &\scalebox{0.78}{0.261} &\scalebox{0.78}{0.381} \\ %&\secondres{\scalebox{0.78}{0.235}} &\secondres{\scalebox{0.78}{0.280}}\\

                \midrule

					\multicolumn{2}{c|}{\scalebox{0.78}{{$1^{\text{st}}$ Count}}} & \scalebox{0.78}{\boldres{34}} & \scalebox{0.78}{\boldres{22}} &{\scalebox{0.78}{1}} &{\scalebox{0.78}{6}}  & \scalebox{0.78}{{1}} & \scalebox{0.78}{\secondres{11}} & \scalebox{0.78}{0}& \scalebox{0.78}{0}& \scalebox{0.78}{0}& \scalebox{0.78}{0}& \scalebox{0.78}{2}& \scalebox{0.78}{0}& \scalebox{0.78}{0}& \scalebox{0.78}{0}& \scalebox{0.78}{0}& \scalebox{0.78}{0}& \scalebox{0.78}{0}& \scalebox{0.78}{0}& \scalebox{0.78}{\secondres{3}} & \scalebox{0.78}{{2}} & \scalebox{0.78}{0}& \scalebox{0.78}{0}& \scalebox{0.78}{0}& \scalebox{0.78}{0} \\% & \scalebox{0.78}{0}& \scalebox{0.78}{0}\\
					\bottomrule
				\end{tabular}
		\end{threeparttable}
	}
\end{table*}

\subsection{Experimental Details}

We adhere to the standard approach for dataset partitioning, splitting all datasets into training, validation, and testing sets as detailed in Table~\ref{datasets}. Our model employs the L2 loss function and utilizes the ADAM optimizer for iterative parameter updates. The training process is set to run for 10 epochs, incorporating early stopping as needed. All experiments are executed using PyTorch on a single NVIDIA GeForce RTX 3090.
Algorithm \ref{forecasting_procedure_of_mamba_block} presents the prediction process of FMamba.

\begin{algorithm}
\caption{The Forecasting Procedure of FMamba}
\label{forecasting_procedure_of_mamba_block}

{\bf Input:} $X = \{ \mathbf{x}_{1}, \ldots, \mathbf{x}_{L} \} \in \mathbb{R}^{B \times L \times n}$

{\bf Output:} $Y = \{ \mathbf{x}_{L+1}, \ldots, \mathbf{x}_{L+\tau} \} \in \mathbb{R}^{B \times \tau \times n}$
\begin{algorithmic}[1]
\State $X^\top \in \mathbb{R}^{B \times n \times L} \leftarrow \texttt{Permute}(X) $ 
\State $X_{\rm norm} \leftarrow \texttt{Norm}(X^\top)$
\State $X_{emb} \leftarrow \texttt{Embedding}(X_{\rm norm})$
\For{$i$ \textbf{in} {\rm FMamba Layers}}:
    \State $Y' \leftarrow \texttt{Fast-attention}(X_i)$  \Comment{$X_i$ represents the $i_{\rm th}$ FMamba Layer's input }
    \State $Y' \leftarrow \texttt{Layer-norm}(Y' + X_i)$
    \State $Y' \leftarrow \texttt{Mamba}(Y')$
    \State $Y' \leftarrow \texttt{Layer-norm}(Y' + X_i)$
    \State $Y' \leftarrow Y' + \texttt{MLP-block}(Y')$
    \State $Y' \leftarrow \texttt{Layer-norm}(Y')$
\EndFor
\State $Y' \in \mathbb{R}^{B \times n \times \tau} \leftarrow \texttt{Projector}(Y') $  
\State $Y \in \mathbb{R}^{B \times \tau \times n} \leftarrow \texttt{Permute}(Y')$
\end{algorithmic}
\end{algorithm}

\subsection{Results and Analysis}
The detailed experimental results on eight datasets are presented in Table \ref{full_baseline_results}, with the best results highlighted in red and the second-best results underlined. It illustrates the comparative forecasting performance of FMamba and other baselines across various datasets and tasks. To facilitate reproduction, we also provide the model hyperparameters corresponding to different tasks shown in the Appendix. Through analysis, we can find that FMamba achieves lower MSE and MAE metrics than other baselines on most tasks.

Specifically, compared to S-Mamba based on Mamba, FMamba demonstrates an overall improvement in average MSE across all datasets: \textbf{22.8\%} (0.136$\rightarrow$0.105) in PEMS03, \textbf{2.1\%} (0.096$\rightarrow$0.094) in PEMS04, \textbf{10.2\%} (0.088$\rightarrow$0.079) in PEMS07, \textbf{26.3\%} (0.156$\rightarrow$0.115) in PEMS08, \textbf{1.2\%} (0.171$\rightarrow$0.169) in Electricity, \textbf{17.6\%} (0.415$\rightarrow$0.342) in SML2010, \textbf{2.0\%} (0.252$\rightarrow$0.247) in Weather, and \textbf{12.7\%} (0.244$\rightarrow$0.213) in Solar-Energy. In terms of the average MAE indicator, FMamba performs better on most datasets except for Electricity and Weather (0.267$\rightarrow$0.269, 0.277$\rightarrow$0.293). Compared to iTransformer based on Transformer, FMamba also shows an overall improvement in average MSE across all datasets: \textbf{7.1\%} (0.113$\rightarrow$0.105) in PEMS03, \textbf{15.3\%} (0.111$\rightarrow$0.094) in PEMS04, \textbf{21.8\%} (0.101$\rightarrow$0.079) in PEMS07, \textbf{23.3\%} (0.150$\rightarrow$0.115) in PEMS08, \textbf{5.1\%} (0.178$\rightarrow$0.169) in Electricity, \textbf{17.8\%} (0.416$\rightarrow$0.342) in SML2010, \textbf{4.3\%} (0.258$\rightarrow$0.247) in Weather, and \textbf{8.6\%} (0.233$\rightarrow$0.213) in Solar-Energy. In terms of the average MAE indicator, FMamba performs better on most datasets except for Solar-Energy (0.262$\rightarrow$0.270).

\begin{figure}[ht]
	\centering
	\includegraphics[width=0.48\textwidth]{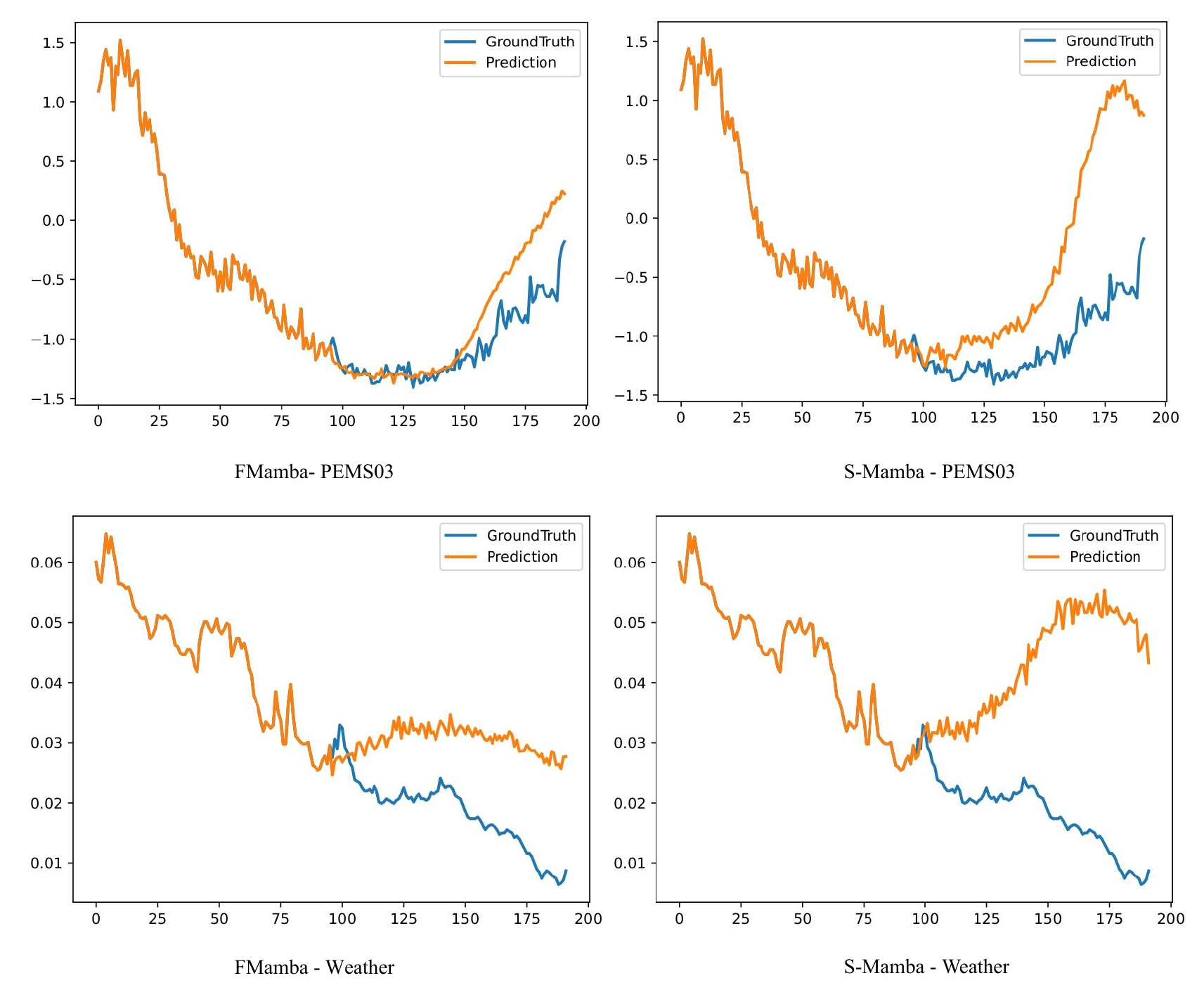}
	\caption{Comparison of forecasting between FMamba and S-Mamba on PEMS03 and Weather when the input length is 96 and the forecasting length is 96.}
	\label{visual-result-partial}
\end{figure}

We attribute FMamba's success to the following two factors: 1) Compared to the bidirectional scanning employed by S-Mamba, FMamba's fast-attention can more comprehensively discern dependencies between variables, aiding the model in learning complex relationships between them; 2) Compared to the traditional Transformer, Mamba not only reduces computational overhead but also enhances the model's robustness by selectively processing or ignoring input information through parameterizing the model's input. To visually demonstrate the prediction performance of FMamba, we visualize its predictions across all datasets and compare them with those of S-Mamba. Part of the visualizations is shown in Figure~\ref{visual-result-partial} and the full visualizations can be found in the Appendix.

\subsection{Ablation Study}

\begin{figure*}[ht]
	\centering
	\includegraphics[width=0.85\textwidth]{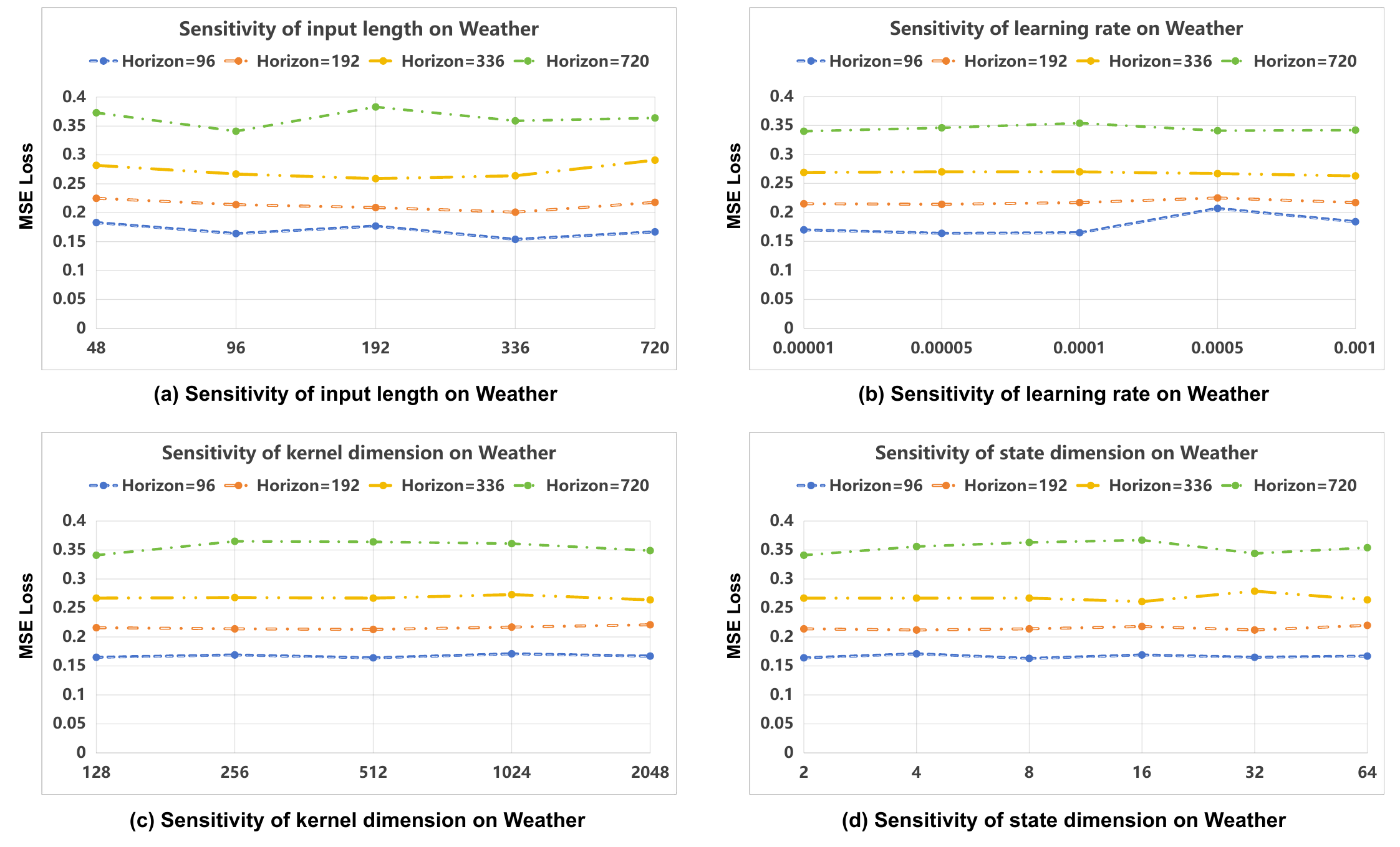}
	\caption{The parameter sensitivity of four components in FMamba.}
	\label{hyper-parameter_sensitivity}
\end{figure*}

To validate the effectiveness of the key components of the model, we design four ablation experiments: 1) To verify whether fast-attention can achieve the same effect as self-attention, we replace fast-attention with self-attention; 2) To compare the impact of Mamba and self-attention on FMamba, we replace Mamba with self-attention; 3) To verify the necessity of fast-attention and Mamba for FMamba, we remove fast-attention and Mamba from FMamba separately. The ablation results are listed in Table~\ref{ablation_simplified} and detailed ablation results are shown in the Appendix. Here are some noteworthy points: 1) Replacing fast-attention with self-attention in FMamba has little impact on the model's predictive performance, indicating that fast-attention can achieve the same effect as canonical self-attention while maintaining linear computational complexity; 2) Replacing Mamba with self-attention in FMamba significantly impacts the model's predictive performance, suggesting that the inclusion of Mamba helps the model selectively process input features, enhancing its robustness; 3) Analyzing the results of the third and fourth ablation experiments, we can see the indispensable role of fast-attention and Mamba in FMamba. This is likely due to the ability of fast-attention to attend to global variations and Mamba's selective processing capability.

\begin{table}[htbp]
\caption{Ablations on FMamba. The average results of all predicted lengths are listed here.}
\label{ablation_simplified}
%\vspace{5pt}
\centering
    \renewcommand{\multirowsetup}{\centering}
    \setlength{\tabcolsep}{2pt}
    \scalebox{0.72}{
    \begin{tabular}{c|c|c|cc|cc|cc}
    \toprule
    \multirow{2}{*}{Design} &\multirow{2}{*}{Cross-variate} & \multirow{2}{*}{Selected} & \multicolumn{2}{c|}{PEMS08} & \multicolumn{2}{c|}{Electricity} & \multicolumn{2}{c}{Solar-Energy} \\
    \cmidrule(lr){4-5} \cmidrule(lr){6-7}  \cmidrule(lr){8-9} 
    &Encoding &Attention & MSE & MAE & MSE & MAE  & MSE & MAE  \\
    \midrule
     \textbf{FMamba} & \textbf{Fast-attention} & \textbf{Mamba} & \boldres{0.115} & \boldres{0.211} & \secondres{0.169} & \secondres{0.269} & 0.213 & 0.270 \\
     
     \midrule
     \multirow{2}{*}{Replace}& Self-attention & \textbf{Mamba} & \secondres{0.116} & \secondres{0.214} & 0.173 & 0.275 & \boldres{0.199} & \boldres{0.264} \\
     & \textbf{Fast-attention} & Self-attention & 0.131 & 0.227 & \boldres{0.166} & \boldres{0.266} & 0.218 & 0.275  \\
     
    \midrule
    \multirow{2}{*}{w/o} & \textbf{Fast-attention} & w/o & 0.124 & 0.215 & 0.170 & 0.270 & 0.213 & 0.271 \\
     & w/o & \textbf{Mamba} & \secondres{0.116} & \secondres{0.214} & 0.180 & 0.278 & \secondres{0.201} & \secondres{0.265}  \\
    \bottomrule
    \end{tabular}
    }
\end{table}

\subsection{Parameter Sensitivity Analysis}
To determine whether FMamba is sensitive to the length of the input sequence and various hyper-parameters, we conduct sensitivity analysis experiments using the Weather dataset. Specifically, we examine the model's Input Length, Learning Rate, Kernel Dimension, and State Dimension. For each parameter setting, we adhere to the hyper-parameters listed in Table~\ref{parameter-E-S-W-S} and keep the other hyper-parameters constant. The detailed results are presented in Figure~\ref{hyper-parameter_sensitivity}.
Overall, the impact of these parameters on the model's predictive performance is minimal, underscoring the robustness of FMamba.

\section{Conclusion}

This paper proposes a novel model for MTSF, named FMamba. It primarily comprises an Embedding layer, fast-attention, Mamba, and an MLP-block. The Embedding layer and the MLP-block of the FMamba layer are respectively utilized to extract lower-level and higher-level temporal features. The fast-attention mechanism helps the model overcome the unilateral nature of Mamba, enabling it to attend to the mutual dependencies of various variates like self-attention, without incurring quadratic computational complexity. The Mamba parameterizes the input features, allowing FMamba to focus on or ignore information between variables selectively. Extensive experiments on various real-world datasets demonstrate that FMamba outperforms SOTA methods in addressing MTSF problems.

\section{Acknowledgments}
This work was supported by the National Natural Science Foundation of China (No. 62173317).

\bibliography{aaai25}

\begin{thebibliography}{21}
\providecommand{\natexlab}[1]{#1}

\bibitem[{Ahamed and Cheng(2024)}]{ahamed2024timemachine}
Ahamed, M.~A.; and Cheng, Q. 2024.
\newblock Timemachine: A time series is worth 4 mambas for long-term
  forecasting.
\newblock \emph{arXiv preprint arXiv:2403.09898}.

\bibitem[{Choromanski et~al.(2021)Choromanski, Likhosherstov, Dohan, Song,
  Gane, Sarl{\'{o}}s, Hawkins, Davis, Mohiuddin, Kaiser, Belanger, Colwell, and
  Weller}]{ChoromanskiLDSG21}
Choromanski, K.~M.; Likhosherstov, V.; Dohan, D.; Song, X.; Gane, A.;
  Sarl{\'{o}}s, T.; Hawkins, P.; Davis, J.~Q.; Mohiuddin, A.; Kaiser, L.;
  Belanger, D.~B.; Colwell, L.~J.; and Weller, A. 2021.
\newblock Rethinking Attention with Performers.
\newblock In \emph{International Conference on Learning Representations}.

\bibitem[{Das et~al.(2023)Das, Kong, Leach, Sen, and Yu}]{das2023long}
Das, A.; Kong, W.; Leach, A.; Sen, R.; and Yu, R. 2023.
\newblock {Long-term Forecasting with TiDE: Time-series Dense Encoder}.
\newblock \emph{arXiv preprint arXiv:2304.08424}.

\bibitem[{Gu and Dao(2023)}]{gu2023mamba}
Gu, A.; and Dao, T. 2023.
\newblock Mamba: Linear-time sequence modeling with selective state spaces.
\newblock \emph{arXiv preprint arXiv:2312.00752}.

\bibitem[{Gu et~al.(2020)Gu, Dao, Ermon, Rudra, and
  R\'{e}}]{NEURIPS2020_102f0bb6}
Gu, A.; Dao, T.; Ermon, S.; Rudra, A.; and R\'{e}, C. 2020.
\newblock {HiPPO: Recurrent Memory with Optimal Polynomial Projections}.
\newblock In \emph{Advances in Neural Information Processing Systems},
  volume~33, 1474--1487.

\bibitem[{Gu, Goel, and R{\'e}(2021)}]{gu2021efficiently}
Gu, A.; Goel, K.; and R{\'e}, C. 2021.
\newblock Efficiently modeling long sequences with structured state spaces.
\newblock \emph{arXiv preprint arXiv:2111.00396}.

\bibitem[{Li et~al.(2023)Li, Qi, Li, and Xu}]{li2023revisiting}
Li, Z.; Qi, S.; Li, Y.; and Xu, Z. 2023.
\newblock {Revisiting long-term time series forecasting: An investigation on
  linear mapping}.
\newblock \emph{arXiv preprint arXiv:2305.10721}.

\bibitem[{LIU et~al.(2022)LIU, Zeng, Chen, Xu, LAI, Ma, and Xu}]{liu2022scinet}
LIU, M.; Zeng, A.; Chen, M.; Xu, Z.; LAI, Q.; Ma, L.; and Xu, Q. 2022.
\newblock SCINet: Time Series Modeling and Forecasting with Sample Convolution
  and Interaction.
\newblock In \emph{Advances in Neural Information Processing Systems},
  volume~35, 5816--5828.

\bibitem[{Liu et~al.(2024)Liu, Hu, Zhang, Wu, Wang, Ma, and
  Long}]{liu2024itransformer}
Liu, Y.; Hu, T.; Zhang, H.; Wu, H.; Wang, S.; Ma, L.; and Long, M. 2024.
\newblock {iTransformer: Inverted Transformers Are Effective for Time Series
  Forecasting}.
\newblock In \emph{International Conference on Learning Representations}.

\bibitem[{Liu et~al.(2022)Liu, Wu, Wang, and Long}]{NEURIPS2022_4054556f}
Liu, Y.; Wu, H.; Wang, J.; and Long, M. 2022.
\newblock {Non-stationary Transformers: Exploring the Stationarity in Time
  Series Forecasting}.
\newblock In \emph{Advances in Neural Information Processing Systems},
  volume~35, 9881--9893.

\bibitem[{Ma et~al.(2023)Ma, Zhang, Zhao, Kang, and Bai}]{ma2023tcln}
Ma, S.; Zhang, T.; Zhao, Y.-B.; Kang, Y.; and Bai, P. 2023.
\newblock {TCLN: A Transformer-based Conv-LSTM network for multivariate time
  series forecasting}.
\newblock \emph{Applied Intelligence}, 53(23): 28401--28417.

\bibitem[{Ma et~al.(2024)Ma, Zhao, Kang, and Bai}]{10552140}
Ma, S.; Zhao, Y.-B.; Kang, Y.; and Bai, P. 2024.
\newblock {Multivariate Time Series Modeling and Forecasting with Parallelized
  Convolution and Decomposed Sparse-Transformer}.
\newblock \emph{IEEE Transactions on Artificial Intelligence}, 1--11.

\bibitem[{Nie et~al.(2023)Nie, H.~Nguyen, Sinthong, and
  Kalagnanam}]{Yuqietal-2023-PatchTST}
Nie, Y.; H.~Nguyen, N.; Sinthong, P.; and Kalagnanam, J. 2023.
\newblock {A Time Series is Worth 64 Words: Long-term Forecasting with
  Transformers}.
\newblock In \emph{International Conference on Learning Representations}.

\bibitem[{Vaswani et~al.(2017)Vaswani, Shazeer, Parmar, Uszkoreit, Jones,
  Gomez, Kaiser, and Polosukhin}]{NIPS2017_3f5ee243}
Vaswani, A.; Shazeer, N.; Parmar, N.; Uszkoreit, J.; Jones, L.; Gomez, A.~N.;
  Kaiser, L.~u.; and Polosukhin, I. 2017.
\newblock {Attention is All you Need}.
\newblock In \emph{Advances in Neural Information Processing Systems},
  volume~30.

\bibitem[{Wang et~al.(2023)Wang, Wang, Peng, Zhang, and Tang}]{wang2023hybrid}
Wang, X.; Wang, Y.; Peng, J.; Zhang, Z.; and Tang, X. 2023.
\newblock A hybrid framework for multivariate long-sequence time series
  forecasting.
\newblock \emph{Applied Intelligence}, 53(11): 13549--13568.

\bibitem[{Wang et~al.(2024)Wang, Kong, Feng, Wang, Zhao, Wang, and
  Zhang}]{wang2024mamba}
Wang, Z.; Kong, F.; Feng, S.; Wang, M.; Zhao, H.; Wang, D.; and Zhang, Y. 2024.
\newblock {Is Mamba Effective for Time Series Forecasting?}
\newblock \emph{arXiv preprint arXiv:2403.11144}.

\bibitem[{Wu et~al.(2023)Wu, Hu, Liu, Zhou, Wang, and Long}]{wu2022timesnet}
Wu, H.; Hu, T.; Liu, Y.; Zhou, H.; Wang, J.; and Long, M. 2023.
\newblock {TimesNet: Temporal 2D-Variation Modeling for General Time Series
  Analysis}.
\newblock In \emph{International Conference on Learning Representations}.

\bibitem[{Zeng et~al.(2023)Zeng, Chen, Zhang, and Xu}]{zeng2023transformers}
Zeng, A.; Chen, M.; Zhang, L.; and Xu, Q. 2023.
\newblock Are transformers effective for time series forecasting?
\newblock In \emph{Proceedings of the AAAI conference on artificial
  intelligence}, volume~37, 11121--11128.

\bibitem[{Zhang and Yan(2023)}]{zhang2022crossformer}
Zhang, Y.; and Yan, J. 2023.
\newblock {Crossformer: Transformer utilizing cross-dimension dependency for
  multivariate time series forecasting}.
\newblock In \emph{International Conference on Learning Representations}.

\bibitem[{Zhou et~al.(2021)Zhou, Zhang, Peng, Zhang, Li, Xiong, and
  Zhang}]{zhou2021informer}
Zhou, H.; Zhang, S.; Peng, J.; Zhang, S.; Li, J.; Xiong, H.; and Zhang, W.
  2021.
\newblock Informer: Beyond efficient transformer for long sequence time-series
  forecasting.
\newblock In \emph{Proceedings of the AAAI conference on artificial
  intelligence}, volume~35, 11106--11115.

\bibitem[{Zhou et~al.(2022)Zhou, Ma, Wen, Wang, Sun, and
  Jin}]{zhou2022fedformer}
Zhou, T.; Ma, Z.; Wen, Q.; Wang, X.; Sun, L.; and Jin, R. 2022.
\newblock {FEDformer: Frequency Enhanced Decomposed Transformer for Long-term
  Series Forecasting}.
\newblock In \emph{International Conference on Machine Learning}, 27268--27286.

\end{thebibliography}

\appendix

\section{Appendix} \label{appendix}

%\subsection{Hyper-parameters of FMamba on diverse tasks}

\begin{table*}[htbp]
	\caption{The hyper-parameters of FMamba on PEMS03, PEMS04, PEMS07, and PEMS08 datasets for MTSF tasks. }
 \label{parameters-pems}
	\centering
	\resizebox{\textwidth}{!}
	{
		\begin{threeparttable}

				\begin{tabular}{c|c|cccc|cccc|cccc|cccc}
					\toprule
					\multicolumn{2}{c|}{\multirow{1}{*}{Models}}  &
					\multicolumn{4}{c|}{\rotatebox{0}{\scalebox{0.8}{PEMS03}}} &
					\multicolumn{4}{c|}{\rotatebox{0}{\scalebox{0.8}{PEMS04}}} &
					\multicolumn{4}{c|}{\rotatebox{0}{\scalebox{0.8}{{PEMS07}}}} &
					\multicolumn{4}{c}{\rotatebox{0}{\scalebox{0.8}{PEMS08}}}   \\
			      \cmidrule(lr){1-2}
					\cmidrule(lr){3-6} \cmidrule(lr){7-10}\cmidrule(lr){11-14} \cmidrule(lr){15-18}
					\multicolumn{2}{c|}{Horizon}  & \scalebox{0.78}{12} & \scalebox{0.78}{24}  & \scalebox{0.78}{48} & \scalebox{0.78}{96}  & \scalebox{0.78}{12} & \scalebox{0.78}{24}  & \scalebox{0.78}{48} & \scalebox{0.78}{96}  & \scalebox{0.78}{12} & \scalebox{0.78}{24}  & \scalebox{0.78}{48} & \scalebox{0.78}{96}  & \scalebox{0.78}{12} & \scalebox{0.78}{24}  & \scalebox{0.78}{48} & \scalebox{0.78}{96}  \\
					\toprule
					
					\multirow{7}{*}{\rotatebox{90}{\scalebox{0.95}{Hyperparameter}}}
					&  \scalebox{0.78}{$el$} & {\scalebox{0.78}{4}} &{\scalebox{0.78}{4}} & \scalebox{0.78}{4} &\scalebox{0.78}{4} & {\scalebox{0.78}{4}} &{\scalebox{0.78}{4}} & \scalebox{0.78}{4} &\scalebox{0.78}{4} & {\scalebox{0.78}{2}} &{\scalebox{0.78}{2}} & \scalebox{0.78}{2} &\scalebox{0.78}{2} & {\scalebox{0.78}{2}} &{\scalebox{0.78}{2}} & \scalebox{0.78}{2} &\scalebox{0.78}{2}    \\
     
					& \scalebox{0.78}{$bs$} & {\scalebox{0.78}{32}} &{\scalebox{0.78}{32}} & \scalebox{0.78}{32} &\scalebox{0.78}{32} & {\scalebox{0.78}{32}} &{\scalebox{0.78}{32}} & \scalebox{0.78}{32} &\scalebox{0.78}{32} & {\scalebox{0.78}{16}} &{\scalebox{0.78}{16}} & \scalebox{0.78}{16} &\scalebox{0.78}{16} & {\scalebox{0.78}{32}} &{\scalebox{0.78}{32}} & \scalebox{0.78}{32} &\scalebox{0.78}{32}     \\
     
					& \scalebox{0.78}{$lr$} & {\scalebox{0.78}{1e-3}} &{\scalebox{0.78}{1e-3}} & \scalebox{0.78}{1e-3} &\scalebox{0.78}{1e-3} & {\scalebox{0.78}{5e-4}} &{\scalebox{0.78}{5e-4}} & \scalebox{0.78}{5e-4} &\scalebox{0.78}{5e-4} & {\scalebox{0.78}{5e-4}} &{\scalebox{0.78}{5e-4}} & \scalebox{0.78}{5e-4} &\scalebox{0.78}{5e-4} & {\scalebox{0.78}{8e-4}} &{\scalebox{0.78}{8e-4}} & \scalebox{0.78}{8e-4} &\scalebox{0.78}{8e-4}    \\
     
					& \scalebox{0.78}{$d\_{\rm model}$} & {\scalebox{0.78}{512}} &{\scalebox{0.78}{512}} & \scalebox{0.78}{512} &\scalebox{0.78}{512} & {\scalebox{0.78}{1024}} &{\scalebox{0.78}{1024}} & \scalebox{0.78}{1024} &\scalebox{0.78}{1024} & {\scalebox{0.78}{512}} &{\scalebox{0.78}{512}} & \scalebox{0.78}{512} &\scalebox{0.78}{512} & {\scalebox{0.78}{512}} &{\scalebox{0.78}{512}} & \scalebox{0.78}{512} &\scalebox{0.78}{512}    \\

					& \scalebox{0.78}{$\rm dropout$} & {\scalebox{0.78}{0.1}} &{\scalebox{0.78}{0.1}} & \scalebox{0.78}{0.1} &\scalebox{0.78}{0.1} & {\scalebox{0.78}{0.1}} &{\scalebox{0.78}{0.1}} & \scalebox{0.78}{0.1} &\scalebox{0.78}{0.1} & {\scalebox{0.78}{0.1}} &{\scalebox{0.78}{0.1}} & \scalebox{0.78}{0.1} &\scalebox{0.78}{0.1} & {\scalebox{0.78}{0.1}} &{\scalebox{0.78}{0.1}} & \scalebox{0.78}{0.1} &\scalebox{0.78}{0.1}    \\

					& \scalebox{0.78}{$d\_{\rm state}$} & {\scalebox{0.78}{2}} &{\scalebox{0.78}{2}} & \scalebox{0.78}{2} &\scalebox{0.78}{2} & {\scalebox{0.78}{2}} &{\scalebox{0.78}{2}} & \scalebox{0.78}{2} &\scalebox{0.78}{2} & {\scalebox{0.78}{2}} &{\scalebox{0.78}{2}} & \scalebox{0.78}{2} &\scalebox{0.78}{2} & {\scalebox{0.78}{2}} &{\scalebox{0.78}{2}} & \scalebox{0.78}{2} &\scalebox{0.78}{2}   \\	

                 & \scalebox{0.78}{$k\_{dim}$} & {\scalebox{0.78}{128}} &{\scalebox{0.78}{128}} & \scalebox{0.78}{128} &\scalebox{0.78}{128} & {\scalebox{0.78}{128}} &{\scalebox{0.78}{128}} & \scalebox{0.78}{128} &\scalebox{0.78}{128} & {\scalebox{0.78}{512}} &{\scalebox{0.78}{512}} & \scalebox{0.78}{512} &\scalebox{0.78}{512} & {\scalebox{0.78}{512}} &{\scalebox{0.78}{512}} & \scalebox{0.78}{512} &\scalebox{0.78}{512}   \\	
					\bottomrule
				\end{tabular}
    \begin{tablenotes}
			\item[1] $el, bs, lr, d_{\rm model}, d\_{\rm state},$ and $k\_{dim}$ denote the number of encoder layers, batch size, learning rate, feature representation dimension, state dimension of Mamba, and kernel dimension of fast-attention respectively.
		\end{tablenotes}
		\end{threeparttable}

	}
\end{table*}

\begin{table*}[h]
	\caption{The hyper-parameters of FMamba on Electricity, SML2010, Weather, and Solar-Energy datasets for MTSF tasks.
	}\label{parameter-E-S-W-S}
	\centering
	\resizebox{\textwidth}{!}
	{
		\begin{threeparttable}
			\begin{tabular}{c|c|cccc|cccc|cccc|cccc}
				\toprule
				\multicolumn{2}{c|}{\multirow{1}{*}{Models}}  &
				\multicolumn{4}{c|}{\rotatebox{0}{\scalebox{0.8}{Electricity}}} &
				\multicolumn{4}{c|}{\rotatebox{0}{\scalebox{0.8}{{SML2010}}}} &
				\multicolumn{4}{c|}{\rotatebox{0}{\scalebox{0.8}{Weather}}}  &
				\multicolumn{4}{c}{\rotatebox{0}{\scalebox{0.8}{Solar-Energy}}}   \\% &\multicolumn{2}{c}{\rotatebox{0}{\scalebox{0.8}{Informer}}} \\
				%& \multicolumn{2}{c}{\scalebox{0.8}{\citeyearpar{Informer}}} \\
				\cmidrule(lr){1-2}
				\cmidrule(lr){3-6} \cmidrule(lr){7-10}\cmidrule(lr){11-14} \cmidrule(lr){15-18} 
				\multicolumn{2}{c|}{Horizon}  & \scalebox{0.78}{96} & \scalebox{0.78}{192}  & \scalebox{0.78}{336} & \scalebox{0.78}{720}  & \scalebox{0.78}{48} & \scalebox{0.78}{96}  & \scalebox{0.78}{192} & \scalebox{0.78}{336} & \scalebox{0.78}{96} & \scalebox{0.78}{192}  & \scalebox{0.78}{336} & \scalebox{0.78}{720} & \scalebox{0.78}{96} & \scalebox{0.78}{192}  & \scalebox{0.78}{336} & \scalebox{0.78}{720}  \\
				\toprule
				
				\multirow{7}{*}{\rotatebox{90}{\scalebox{0.95}{Hyperparameter}}}
				&  \scalebox{0.78}{$el$} & {\scalebox{0.78}{3}} &{\scalebox{0.78}{3}} & \scalebox{0.78}{3} &\scalebox{0.78}{3} & {\scalebox{0.78}{3}} &{\scalebox{0.78}{3}} & \scalebox{0.78}{3} &\scalebox{0.78}{3} & {\scalebox{0.78}{3}} &{\scalebox{0.78}{3}} & \scalebox{0.78}{3} &\scalebox{0.78}{3} & {\scalebox{0.78}{2}} &{\scalebox{0.78}{2}} & \scalebox{0.78}{2} &\scalebox{0.78}{2}   \\% &\scalebox{0.78}{0.126} &\scalebox{0.78}{0.233}\\
    
				& \scalebox{0.78}{$bs$} & {\scalebox{0.78}{16}} &{\scalebox{0.78}{16}} & \scalebox{0.78}{16} &\scalebox{0.78}{16} & {\scalebox{0.78}{32}} &{\scalebox{0.78}{32}} & \scalebox{0.78}{32} &\scalebox{0.78}{32} & {\scalebox{0.78}{16}} &{\scalebox{0.78}{16}} & \scalebox{0.78}{16} &\scalebox{0.78}{16} & {\scalebox{0.78}{32}} &{\scalebox{0.78}{32}} & \scalebox{0.78}{32} &\scalebox{0.78}{32}    \\%&\scalebox{0.78}{0.139} &\scalebox{0.78}{0.250}\\
    
				& \scalebox{0.78}{$lr$} & {\scalebox{0.78}{8e-4}} &{\scalebox{0.78}{1e-3}} & \scalebox{0.78}{1e-3} &\scalebox{0.78}{1e-3} & {\scalebox{0.78}{8e-4}} &{\scalebox{0.78}{8e-4}} & \scalebox{0.78}{8e-4} &\scalebox{0.78}{8e-4} & {\scalebox{0.78}{5e-5}} &{\scalebox{0.78}{5e-5}} & \scalebox{0.78}{7e-5} &\scalebox{0.78}{7e-5} & {\scalebox{0.78}{1e-4}} &{\scalebox{0.78}{1e-4}} & \scalebox{0.78}{1e-4} &\scalebox{0.78}{1e-4}  \\% &\scalebox{0.78}{0.186} &\scalebox{0.78}{0.289}\\
    
				& \scalebox{0.78}{$d\_{\rm model}$} & {\scalebox{0.78}{512}} &{\scalebox{0.78}{512}} & \scalebox{0.78}{512} &\scalebox{0.78}{512} & {\scalebox{0.78}{512}} &{\scalebox{0.78}{512}} & \scalebox{0.78}{512} &\scalebox{0.78}{512} & {\scalebox{0.78}{512}} &{\scalebox{0.78}{512}} & \scalebox{0.78}{512} &\scalebox{0.78}{512} & {\scalebox{0.78}{512}} &{\scalebox{0.78}{512}} & \scalebox{0.78}{512} &\scalebox{0.78}{512}  \\% &\scalebox{0.78}{0.233} &\scalebox{0.78}{0.323}\\
				
				& \scalebox{0.78}{$\rm dropout$} & {\scalebox{0.78}{0.1}} &{\scalebox{0.78}{0.1}} & \scalebox{0.78}{0.1} &\scalebox{0.78}{0.1} & {\scalebox{0.78}{0.1}} &{\scalebox{0.78}{0.1}} & \scalebox{0.78}{0.1} &\scalebox{0.78}{0.1} & {\scalebox{0.78}{0.1}} &{\scalebox{0.78}{0.1}} & \scalebox{0.78}{0.1} &\scalebox{0.78}{0.1} & {\scalebox{0.78}{0.1}} &{\scalebox{0.78}{0.1}} & \scalebox{0.78}{0.1} &\scalebox{0.78}{0.1}  \\% &\scalebox{0.78}{0.233} &\scalebox{0.78}{0.323}\\
				
				& \scalebox{0.78}{$d\_{\rm state}$} & {\scalebox{0.78}{16}} &{\scalebox{0.78}{16}} & \scalebox{0.78}{16} &\scalebox{0.78}{16} & {\scalebox{0.78}{2}} &{\scalebox{0.78}{2}} & \scalebox{0.78}{2} &\scalebox{0.78}{2} & {\scalebox{0.78}{2}} &{\scalebox{0.78}{2}} & \scalebox{0.78}{2} &\scalebox{0.78}{2} & {\scalebox{0.78}{2}} &{\scalebox{0.78}{2}} & \scalebox{0.78}{2} &\scalebox{0.78}{2}  \\% &\scalebox{0.78}{0.233} &\scalebox{0.78}{0.323}\\

                & \scalebox{0.78}{$k\_{dim}$} & {\scalebox{0.78}{512}} &{\scalebox{0.78}{512}} & \scalebox{0.78}{512} &\scalebox{0.78}{512} & {\scalebox{0.78}{128}} &{\scalebox{0.78}{128}} & \scalebox{0.78}{128} &\scalebox{0.78}{256} & {\scalebox{0.78}{512}} &{\scalebox{0.78}{256}} & \scalebox{0.78}{128} &\scalebox{0.78}{128} & {\scalebox{0.78}{256}} &{\scalebox{0.78}{256}} & \scalebox{0.78}{256} &\scalebox{0.78}{256}   \\
				
				\bottomrule
			\end{tabular}
		\end{threeparttable}
	}
\end{table*}

\begin{figure*}[ht]
	\centering
	\includegraphics[width=\textwidth]{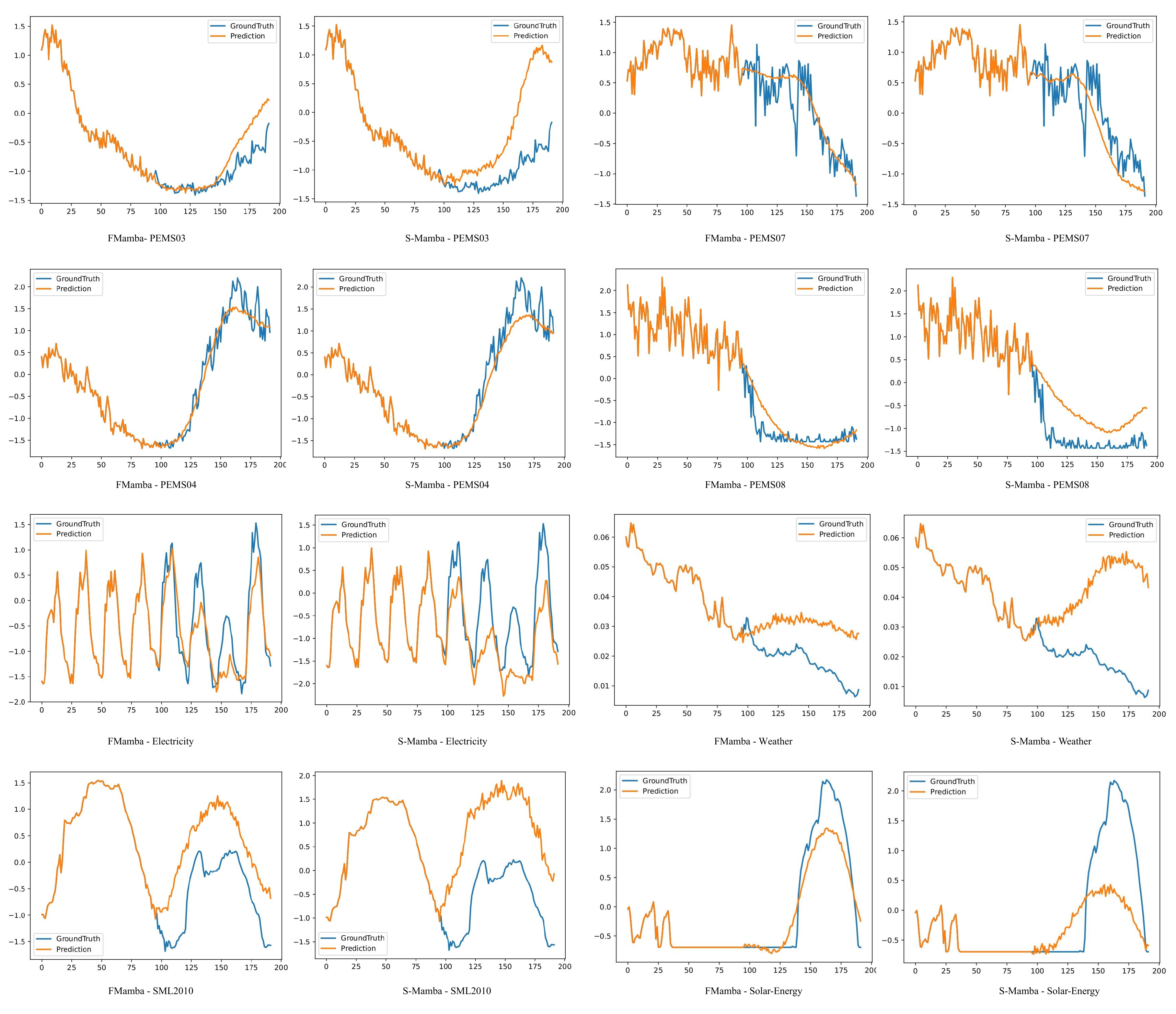}
	\caption{Comparison of forecasts between FMamba and S-Mamba on eight datasets when the input length is 96 and the forecast length is 96. The blue line represents the ground truth and the orange line represents the forecast.}
	\label{visual-result}
\end{figure*}

%\subsection{Full results of the ablation on FMamba}

\begin{table*}[htbp]
\caption{Full results of the ablation on FMamba. We replace different components on the respective dimensions and remove the specific component of FMamba.}
\vspace{3pt}
\label{ablation}
\centering
%\resizebox{1\columnwidth}{!}{
\begin{small}
    \renewcommand{\multirowsetup}{\centering}
    \setlength{\tabcolsep}{3.0pt}
    \begin{tabular}{c|c|c|c|cc|c|cc|cc}
    \toprule
    \multirow{2}{*}{Design} &\multirow{2}{*}{Cross-variate} & \multirow{2}{*}{Selected} & Forecasting  & \multicolumn{2}{c|}{PEMS08} & Forecasting & \multicolumn{2}{c|}{Electricity} & \multicolumn{2}{c}{Solar-Energy}\\
    \cmidrule(lr){5-6}  \cmidrule(lr){8-9} \cmidrule(lr){10-11} 
    &Encoding &Attention &Lengths & MSE & MAE &Lengths  & MSE & MAE & MSE & MAE \\
    
    \midrule
    \multirow{5}{*}{\textbf{FMamba}}& \multirow{5}{*}{\textbf{Fast-attention}} & \multirow{5}{*}{\textbf{Mamba}} & 12 & 0.072 & 0.171 & 96 & 0.137 & 0.237 & 0.195 & 0.251\\
    & & & 24 & 0.091 & 0.192 & 192 & 0.157 & 0.255 & 0.219 & 0.275\\
    & & & 48 & 0.121 & 0.221 & 336 & 0.174 & 0.276 & 0.218 & 0.276\\
    & & & 96 & 0.177 & 0.260 & 720 & 0.208 & 0.308 & 0.221 & 0.279\\
    \cmidrule(lr){4-11}
     & & & Avg & \boldres{0.115} & \boldres{0.211} & Avg & \secondres{0.169} & \secondres{0.269} & {0.213} & {0.270}\\
     \midrule

     \multirow{10}{*}{Replace}& \multirow{5}{*}{Self-attention} & \multirow{5}{*}{\textbf{Mamba}} & 12 & 0.071 & 0.170 & 96 & 0.140 & 0.241 & 0.177 & 0.239\\
     & & & 24 & 0.092 & 0.194 & 192 & 0.162 & 0.264 & 0.196 & 0.263\\
     & & & 48 & 0.122 & 0.223 & 336 & 0.183 & 0.286 & 0.210 & 0.273\\
     & & & 96 & 0.180 & 0.270 & 720 & 0.207 & 0.310 & 0.212 & 0.279\\
     \cmidrule(lr){4-11}
     & & & Avg & \secondres{0.116} & \secondres{0.214} & Avg & 0.173 & 0.275 & \boldres{0.199} & \boldres{0.264}\\
     \cmidrule(lr){2-11}
     
     & \multirow{5}{*}{\textbf{Fast-attention}} & \multirow{5}{*}{Self-attention} & 12 & 0.073 & 0.176 & 96 & 0.139 & 0.237 & 0.199 & 0.257\\
     & & & 24 & 0.095 & 0.202 & 192 & 0.155 & 0.254 & 0.221 & 0.279 \\
     & & & 48 & 0.136 & 0.240 & 336 & 0.170 & 0.274 & 0.225 & 0.283\\
     & & & 96 & 0.219 & 0.290 & 720 & 0.199 & 0.300 & 0.225 & 0.281\\
     \cmidrule(lr){4-11}
     & & & Avg & 0.131 & 0.227 & Avg & \boldres{0.166} & \boldres{0.266} & 0.218 & 0.275 \\

    \midrule

    \multirow{10}{*}{w/o} & \multirow{5}{*}{\textbf{Fast-attention}} & \multirow{5}{*}{w/o} & 12 & 0.071 & 0.171 & 96 & 0.140 & 0.240 & 0.195 & 0.253\\
    & & & 24 & 0.097 & 0.193 & 192 & 0.154 & 0.253 & 0.212 & 0.271\\
     & & & 48 & 0.127 & 0.225 & 336 & 0.172 & 0.275 & 0.223 & 0.280\\
     & & & 96 & 0.199 & 0.269 & 720 & 0.214 & 0.313 & 0.222 & 0.279\\
     \cmidrule(lr){4-11}
     & & & Avg & 0.124 & 0.215 & Avg & 0.170 & 0.270 & \secondres{0.213} & 0.271\\
     \cmidrule(lr){2-11}
     
     & \multirow{5}{*}{w/o} & \multirow{4}{*}{\textbf{Mamba}} & 12 & 0.073 & 0.174 & 96 & 0.147 & 0.247 & 0.176 & 0.242\\
     & & & 24 & 0.092 & 0.194 & 192 & 0.166 & 0.263 & 0.200 & 0.265\\
     & & & 48 & 0.126 & 0.227 & 336 & 0.185 & 0.284 & 0.214 & 0.276\\
     & & & 96 & 0.171 & 0.262 & 720 & 0.222 & 0.318 & 0.212 & 0.276\\
     \cmidrule(lr){4-11}
     & & & Avg & \secondres{0.116} & \secondres{0.214} & Avg & 0.180 & 0.278 & \secondres{0.201} & \secondres{0.265}\\
    \bottomrule
    \end{tabular}
\end{small}
%}
\end{table*}

\end{document}